\documentclass{article}

 \PassOptionsToPackage{numbers, compress}{natbib}

\usepackage[final]{neurips_2025}

\usepackage[utf8]{inputenc} 
\usepackage[T1]{fontenc}    
\usepackage{hyperref}       
\usepackage{url}            
\usepackage{booktabs}       
\usepackage{amsfonts}       
\usepackage{nicefrac}       
\usepackage{microtype}      
\usepackage{xcolor}         
\usepackage[table,xcdraw]{xcolor}

\usepackage{hyperref}
\usepackage{CJKutf8}
\usepackage{graphicx} 
\usepackage{amsmath}
\usepackage{tikz}
\usepackage{multirow}
\usepackage{amssymb}
\usepackage[normalem]{ulem}
\usepackage{esvect} 

\usepackage{graphicx} 
\usepackage{lipsum}   
\usepackage{wrapfig}  

\useunder{\uline}{\ul}{}

\title{

Learning to Insert for Constructive Neural Vehicle Routing Solver

}

\author{
  Fu Luo$^{1,2}$, Xi Lin$^{3}$, Mengyuan Zhong$^4$, Fei Liu$^{3}$, Zhenkun Wang$^{1,2}$\thanks{Corresponding author}\;\,, \textbf{Jianyong Sun$^{4*}$}, \textbf{Qingfu Zhang$^{3}$}  \\
$^1$ School of Automation and Intelligent Manufacturing, \\ Southern University of Science and Technology, Shenzhen, China \\
$^2$ Guangdong Provincial Key Laboratory of Fully Actuated System Control Theory and Technology, \\ Southern University of Science and Technology, Shenzhen, China\\
    $^3$ Department of Computer Science, City University of Hong Kong, Hong Kong SAR, China\\
    $^4$ School of Mathematics and Statistics, Xi’an Jiaotong University, Xi’an, China \\
  \texttt{luof2023@mail.sustech.edu.cn, xi.lin@my.cityu.edu.hk, my.zhong@stu.xjtu.edu.cn,}\\ \texttt{fliu36-c@my.cityu.edu.hk, wangzhenkun90@gmail.com,}\\
  \texttt{jy.sun@xjtu.edu.cn, qingfu.zhang@cityu.edu.hk} \\
  }

\begin{document}
\begin{CJK*}{UTF8}{gbsn}

\maketitle

\begin{abstract}

Neural Combinatorial Optimisation (NCO) is a promising learning-based approach for solving Vehicle Routing Problems (VRPs) without extensive manual design. While existing constructive NCO methods typically follow an appending-based paradigm that sequentially adds unvisited nodes to partial solutions, this rigid approach often leads to suboptimal results. To overcome this limitation, we explore the idea of the insertion-based paradigm and propose Learning to Construct with Insertion-based Paradigm (L2C-Insert), a novel learning-based method for constructive NCO. Unlike traditional approaches, L2C-Insert builds solutions by strategically inserting unvisited nodes at any valid position in the current partial solution, which can significantly enhance the flexibility and solution quality. The proposed framework introduces three key components: a novel model architecture for precise insertion position prediction, an efficient training scheme for model optimization, and an advanced inference technique that fully exploits the insertion paradigm's flexibility. Extensive experiments on both synthetic and real-world instances of the Travelling Salesman Problem (TSP) and Capacitated Vehicle Routing Problem (CVRP) demonstrate that L2C-Insert consistently achieves superior performance across various problem sizes. The code is available at \href{https://github.com/CIAM-Group/L2C\_Insert}{https://github.com/CIAM-Group/L2C\_Insert}.

\end{abstract}

\section{Introduction}

The vehicle routing problem (VRP)~\citep{golden2008vehicle} is an important combinatorial optimization problem with extensive real-world applications in domains such as transportation~\citep{pillac2013review}, logistics~\citep{konstantakopoulos2022vehicle}, and circuit design~\citep{brophy2014principles}. Due to its NP-hard nature~\citep{ausiello2012complexity}, obtaining exact VRP solutions is computationally difficult~\citep{ausiello2012complexity}. The existing approaches, therefore, focus on finding approximate solutions within a practical runtime using problem-specific heuristics. However, designing these heuristics requires substantial domain expertise. Recently, Neural Combinatorial Optimization (NCO) methods have emerged as a promising alternative, as they can leverage neural networks to automatically learn effective heuristics from data, thereby significantly reducing the need for extensive manual design~\citep{vinyals2015pointer,bello2016neural,gao2022multi,zheng2024Pareto_Improver,Sun_Zheng_Wang_2024,zheng2025monte,zhang2025adversarial,li2025multi,li2025ars,zhou2025urs,zhou2025learning,li2025cada,luo2025rethink_tightness}.

Among NCO methods, constructive approaches attract considerable research interest~\citep{vinyals2015pointer,kool2018attention,bengio2021machine,kwon2020pomo,luo2023neural,drakulic2023bq,wang2024distance,zhou2024instance,luo2025boosting}. These methods typically employ an appending-based constructive paradigm that sequentially selects unvisited nodes and appends them to the end of the current partial solution. This paradigm is inspired by the sequence-to-sequence framework~\citep{sutskever2014sequence} in natural language processing (NLP), where the model encodes an input sequence into an embedding and then autoregressively generates an output sequence based on this representation. \citet{vinyals2015pointer} pioneer this adaptation through their Pointer Networks (Ptr-Nets), which establishes the first sequence-to-sequence framework specifically for neural combinatorial optimization. Subsequent research has further advanced constructive NCO methods within this appending-based paradigm, with improvements focusing on model training techniques~\citep{bello2016neural} and architectural enhancements~\citep{kool2018attention}. However, appending-based methods suffer from a fundamental limitation due to their strict operational constraint that only permits adding new nodes to the end of the current partial solution. This constraint prevents potentially more effective modifications and often leads to suboptimal partial solutions caused by early greedy decisions.

\begin{figure*}[t]
\centering
\includegraphics[width=1.0\textwidth]{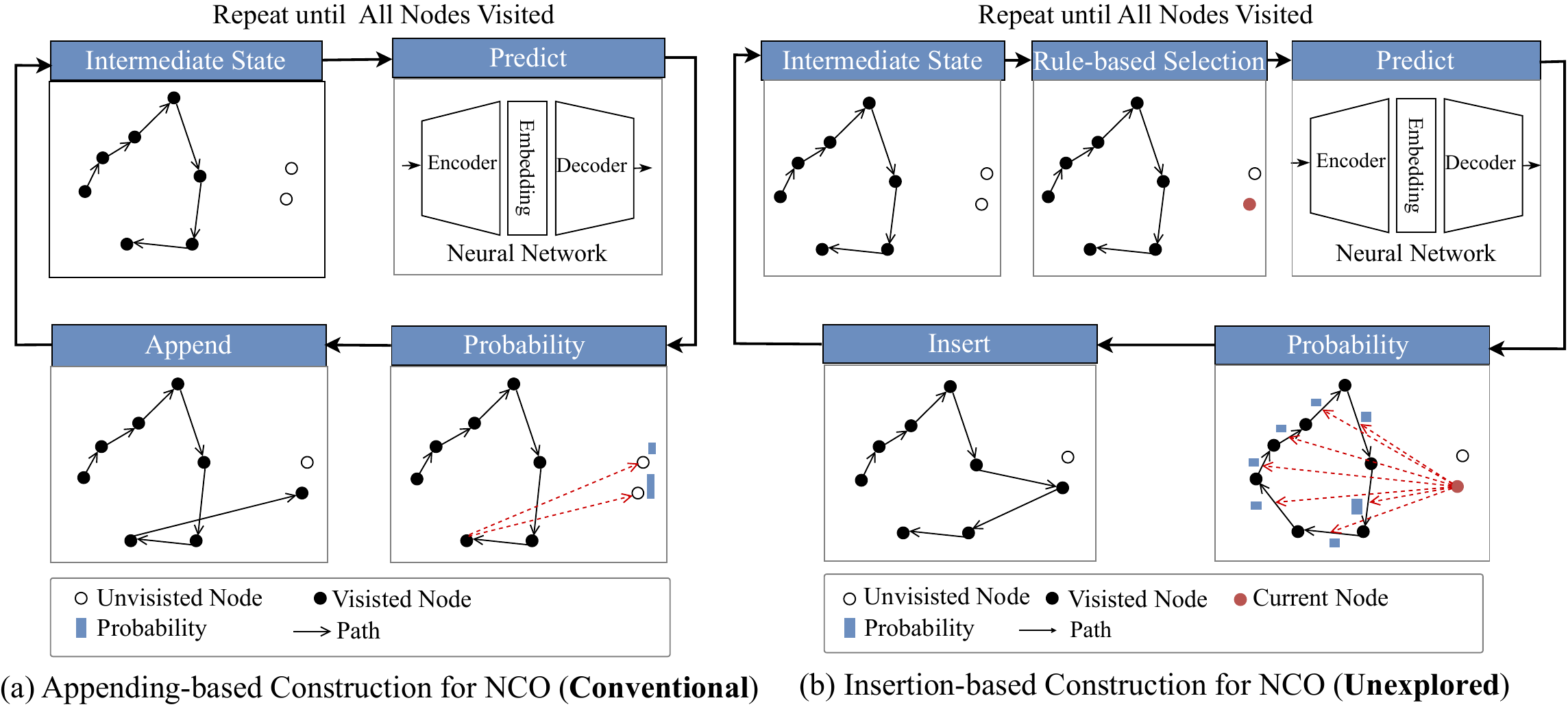}
\caption{
\textbf{Appending-based construction vs. Insertion-based construction.}
Given the same intermediate state in the construction process: \textbf{(a)} The appending-based paradigm restricts the addition of a selected node exclusively to the end of the current partial solution. This inflexibility leads to path intersections, a common cause of suboptimal solutions. \textbf{(b)} In contrast, the insertion-based construction offers greater flexibility by allowing the selected node to be inserted into any feasible position within the partial solution. Such flexibility is crucial for avoiding the intersections and thereby enhancing the potential for generating high-quality solutions.
}
\label{Intro: Append vs. insert}
\end{figure*}

The insertion-based paradigm offers a more effective alternative for addressing VRPs~\citep{rosenkrantz1977analysis,liu2023heuristics}. This approach constructs solutions by sequentially inserting an unvisited node into any valid position within the current partial solution, which potentially offers greater flexibility and yields higher-quality solutions than the appending-based paradigm. For example, as shown in Figure~\ref{Intro: Append vs. insert}, given the same intermediate state in the construction process, the rigid sequential nature of the appending paradigm can only add the selected node to the end of the solution, leading to path intersections and a suboptimal solution. In contrast, the insertion-based construction allows for the flexible placement of the selected node into any feasible position within the partial solution path, thereby avoiding such intersections and holding the potential to generate higher-quality solutions. Despite these clearly demonstrated benefits, the insertion-based constructive approach has remained unexplored in NCO for VRPs in the past decade. A comprehensive review of constructive NCO methods can be found in Appendix~\ref{Appendix: Related Work}.

To bridge this research gap, this work proposes  Learning to Construct with Insertion-based Paradigm (L2C-Insert), a novel insertion-based learning framework for constructive NCO. To the best of our knowledge, L2C-Insert represents the first learning-driven, insertion-based framework specifically designed for solution construction in NCO. Our main contributions can be summarized as follows:
\begin{itemize}
    \item We propose a novel model architecture, an effective training scheme, and a powerful inference technique for the L2C-Insert framework. Together, these components are crucial for enabling the L2C-Insert framework to achieve robust performance.

    \item We develop a supervised learning based training scheme tailored for the efficient learning of the L2C-Insert model. Building on the inherent flexibility of the insertion paradigm, we further propose an insertion-based local reconstruction mechanism to enhance model performance during inference.
    
    \item We conduct comprehensive experiments to explore the advantages of the L2C-Insert framework on both synthetic and real-world benchmarks. The results demonstrate that our L2C-Insert method achieves outstanding performance on TSP and CVRP instances with sizes ranging from 100 to 100K nodes.

\end{itemize}

\section{Preliminaries}
\label{Preliminaries}

\subsection{VRP Formulation}

A VRP instance $S$ can be represented as a graph containing $n$ nodes, where each node $i\in \{1,\ldots,n\}$ is represented by a feature vector $\boldsymbol{s}_i \in \mathbb{R}^{d_s}$. For example, in the TSP, $\boldsymbol{s}_i$ typically corresponds to the 2D coordinates of node $i$. A solution to a VRP instance specifies the sequence, or set of sequences (routes/tours), in which nodes are visited. For example, in the TSP case, a solution is a single tour that can be represented as a permutation of all $n$ nodes: $\boldsymbol{\pi}=(\pi_1,  \ldots,  \pi_t, \ldots, \pi_n)$, where each element $\pi_t \in \{1,\ldots,n\}$ is a distinct node. The objective in solving VRPs is generally to minimize the cost of the solution, which often corresponds to the total Euclidean length of the tour(s). A VRP solution is considered feasible if it satisfies all problem-specific constraints. For example, in the TSP, a feasible solution must visit each node in the graph exactly once. Constraints in the CVRP additionally include vehicle capacity limits, which are further detailed in Appendix~\ref{Implemetation details on CVRP}.

\subsection{Appending-based Construction for Solving VRPs} 

The appending-based constructive heuristic is a prominent approach that NCO methods frequently leverage for solving VRPs~\citep{vinyals2015pointer,kool2018attention,kwon2020pomo,luo2023neural,drakulic2023bq,gao2024towards,pirnay2024selfimprovement,zheng2024udc,hottung2025polynet}. Initializing with an empty solution, this heuristic constructs the solution to a given instance by sequentially appending unvisited nodes to the end of the current partial solution until all nodes are visited. At each construction step, it selects a node from the set of unvisited nodes to append, based on a specified strategy. A common selection strategy is to choose the unvisited node that has the smallest Euclidean distance to the last node in the current partial solution. However, this simple selection strategy typically leads to solutions with relatively low quality. The existing learning-based constructive NCO methods focus on using neural networks to learn more effective selection strategies capable of generating near-optimal solutions.

\subsection{Insertion-based Construction for Solving VRPs}

The insertion-based constructive heuristic, though well-established in Operations Research for solving VRPs~\citep{renaud2000heuristic,hassin2008greedy,liu2023heuristics}, has not been explored for constructive NCO. Similar to the appending-based approach, this heuristic begins with an empty solution and iteratively incorporates unvisited nodes into the partial route until all nodes are included. Unlike simple appending, this method permits insertion of unvisited nodes at any valid position within the current partial solution. Each construction step involves two critical decisions: first, selecting an unvisited node for insertion, and second, identifying the proper position within the current partial solution for insertion. Typical node selection strategies include random selection from the unvisited set or choosing the node closest to the previously inserted node. Regarding position selection, a common heuristic is to choose the insertion node that minimizes the increase in the total solution cost (e.g., tour length) upon inserting the selected node. Our work mainly focuses on leveraging neural networks to learn more effective position selection strategies to generate high-quality solutions.

\begin{figure*}[t]
\centering
\includegraphics[width=0.95\textwidth]{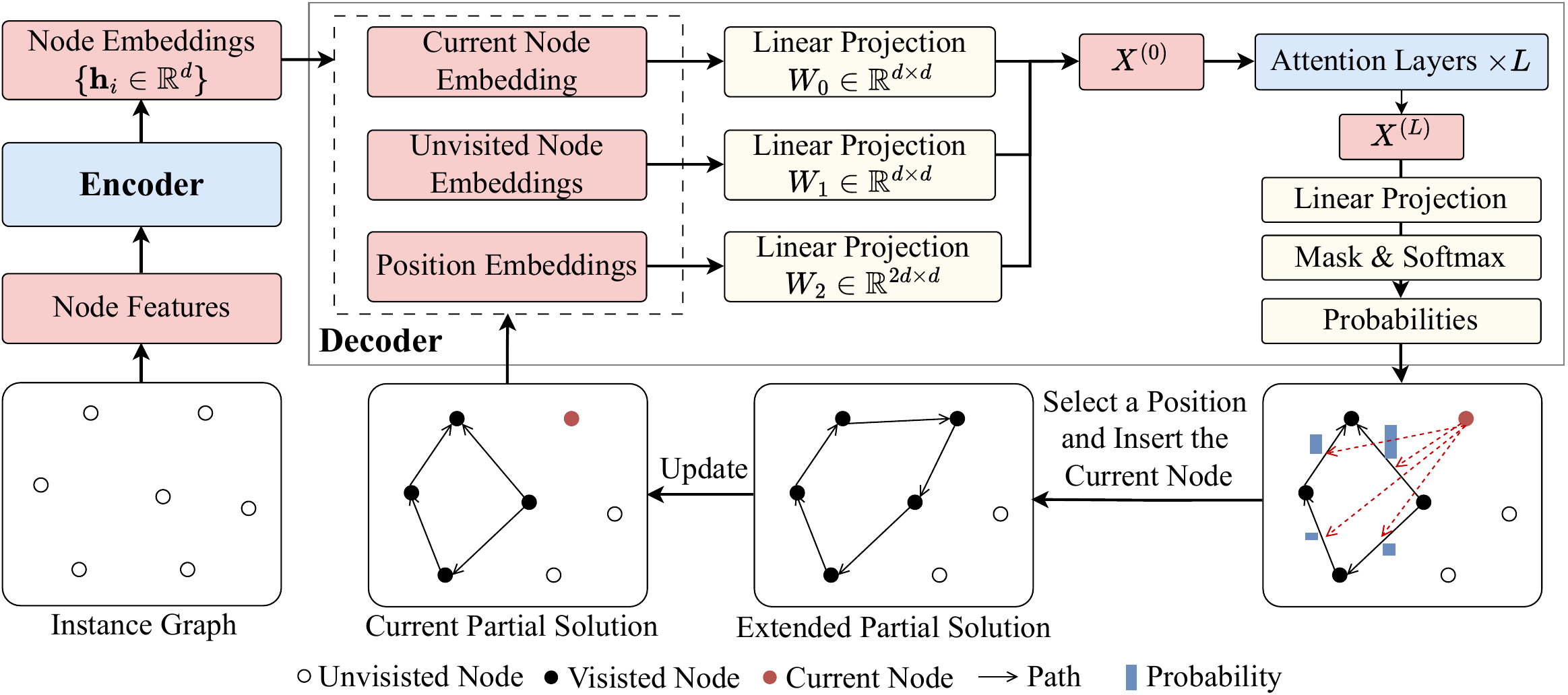}
\caption{\textbf{Model Structure of L2C-Insert.} \textbf{(a) Encoding:} At the beginning of solution construction, the encoder transforms each node's feature vector into a $d$-dimensional node embedding $\mathbf{h}_i\in \mathbb{R}^{d}$. \textbf{(b) Decoding:} At construction step $t$, the decoder maintains three embeddings for 1) the current node (the node to be inserted), 2) unvisited nodes, and 3) the positions in the partial solution, where each position embedding is created by horizontally concatenating the embeddings of adjacent visited nodes. Each embedding type undergoes separate linear transformations before being collectively processed through $L$ stacked attention layers. The final insertion probabilities are obtained by applying a linear projection followed by a masked softmax operation to the refined embeddings.}
\label{Model Structure}
\end{figure*}

\section{L2C-Insert: Learning to Construct in Insertion Paradigm}
\label{Section: Method}

This section proposes the L2C-Insert framework for learning the insertion-based construction heuristic, which integrates three key components: (1) a novel model architecture, (2) an effective training scheme, and (3) a powerful inference technique. We demonstrate this framework through solving TSP, and the required adaptations for solving CVRP are provided in Appendix~\ref{Implementation Details of L2C-Insert}.

\subsection{Model Structure}

Similar to the appending-based constructive NCO models, our L2C-Insert model utilizes an encoder-decoder framework, with the decoder features with multiple attention layers~\citep{vaswani2017attention} for the promising learning ability~\citep{luo2023neural,drakulic2023bq} as shown in Figure~\ref{Model Structure}.

\paragraph{Encoder} The encoder contains a linear projection layer followed by an attention layer. For a problem instance $S$ with $n$ nodes, each node's feature vector $\boldsymbol{s}_i\in \mathbb{R}^{d_s}$ (where $i=1,\dots,n$) is first transformed into an initial node embedding $\mathbf{h}_i^0\in \mathbb{R}^{d}$, where $d$ denotes the embedding dimension. These initial embeddings are subsequently updated through the attention layer~\citep{kwon2020pomo,luo2023neural} to produce the refined embeddings $\mathbf{h}_i\in \mathbb{R}^{d}$, which is the output of the encoder.

\paragraph{Decoder} At construction step $t$, the decoder receives the embeddings of all nodes, denoted as $\mathbf{h}_i\in \mathbb{R}^{d}, i=1,\dots,n$, the current partial solution $(\pi_1, \pi_2, \ldots, \pi_{t-1})$, and the set of unvisited nodes $\{u_j, u_j \notin \{\pi_{1:t-1}\} \}$. To compute the probability of insertion at the current step, embeddings for three distinct inputs are required: 1) the current node (the node to be inserted), 2) unvisited nodes, and 3) the positions in the partial solution.

First of all, the unvisited node with the minimum Euclidean distance to the previously inserted node is selected as the current node to be inserted, of which the embedding is denoted as $\mathbf{h}_{u_c}$. Secondly, the embeddings of the remaining unvisited nodes can be denoted as $H_u=\{\mathbf{h}_{u_j}\}, u_j \notin \{\pi_{1:t-1}\}, u_j \neq u_c$. Two learnable linear projections $W_0\in \mathbb{R}^{d\times d}$ and $W_1\in \mathbb{R}^{d\times d}$ are applied to the current node's embedding and each unvisited node's embedding, respectively. Thirdly, to generate the embedding of a potential insertion position within the partial solution, the embeddings of adjacent visited nodes are horizontally concatenated. Given the set of visited node embeddings $ \{\mathbf{h}_{\pi_i}\}$, the embedding for the insertion position in the solution segment $(\pi_i,\pi_{i+1})$ is computed as:
\begin{equation}
\begin{aligned}
       \mathbf{e}_{\pi_i} &=
      W_2\begin{cases}
		[\mathbf{h}_{\pi_i}, \mathbf{h}_{\pi_{i+1}}] &    i \in \{1: t-2\}, \\
        [\mathbf{h}_{\pi_{t-1}}, \mathbf{h}_{\pi_1}],  &    i=t-1,
      \end{cases}
\end{aligned}
\end{equation}
where $[\cdot,\cdot]$ represents the horizontal concatenation operator. This operator produces an initial $2d$-dimensional initial insertion position embedding, which is subsequently projected to a $d$-dimensional vector via the trainable matrix $W_2\in \mathbb{R}^{2d\times d}$ to maintain dimensional consistency with node embeddings. The set of position embeddings is denoted as $H_e = \{ \mathbf{e}_{\pi_i}\}, i\in\{1:t-1\}$.

Then, these embeddings together are refined by $L$ stacked attention layers. The first layer receives input $X^{(0)}=\{\mathbf{h}_{c}, H_e, H_u\}$, with each subsequent $j$-th attention layer processing the output from the $(j-1)$-th layer, culminating in the final output $X^{(L)}$. Then, a linear projection and softmax function are applied to $X^{(L)}$ to compute the insertion probabilities of the current node into all valid positions. The masking operation assigns $-\infty$ to invalid positions (those corresponding to the current node and non-visited nodes) to ensure proper probability normalization over valid insertion locations. The inserting probability $p_i$ is calculated as:
\begin{equation}
\begin{aligned}
      x_i  & = X^{(L)}_i W_f + b_f,\\
        a_{i} &= \begin{cases}
		x_i &  \forall  i \in \{\pi_{1:t-1}\}, \\
        -\infty & \text{otherwise,} 
    \end{cases} \\
    \mathbf{p} &= \mbox{Softmax}(\mathbf{a}),
\end{aligned}
\end{equation}
where $W_f \in \mathbb{R}^{d\times 1}$ and $ b_f \in \mathbb{R}^{1}$ are learnable parameters. Each $p_i\in \mathbf{p}$ denotes the probability of inserting the current node between consecutive nodes  $(\pi_i, \pi_{i+1})$ in the partial solution. The decoder executes this sampling-and-insertion procedure $n$ times to construct the complete solution $\{\pi_1,\cdots,\pi_n\}$.

\begin{figure}[t]
\centering
\includegraphics[width=0.95\textwidth]{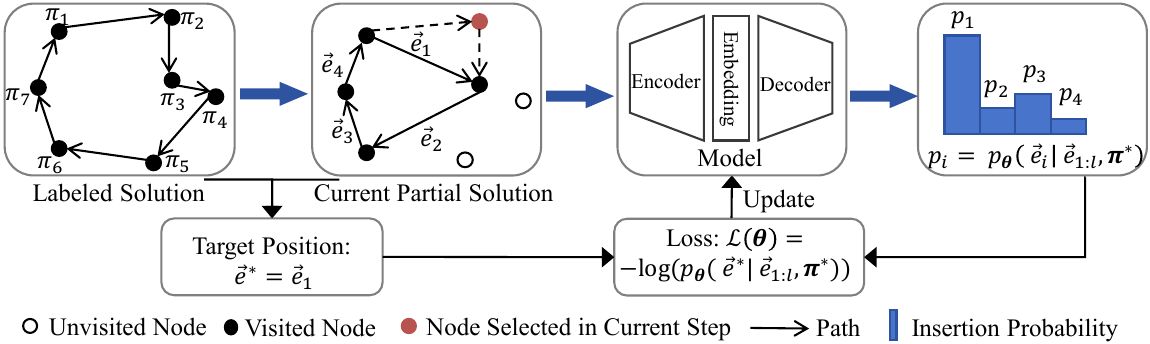}
\caption{\textbf{Training Scheme for L2C-Insert Model.} The model learns to sequentially insert the current node into the target position in the partial solution. The target position for the current node is determined by its adjacent visited nodes in the original labeled solution. The training objective is to maximize the probability of the model predicting the correct target insertion position, as defined by the provided loss function.}
\label{Training example}
\end{figure}

\subsection{Training Scheme}

Constructive NCO models typically adopt either reinforcement learning (RL) or supervised learning (SL) methods for training. The SL-based method can effectively train the model with a heavy decoder, whereas the RL-based method is not capable, as indicated by~\citet{luo2023neural}. Since our model utilizes an insertion-based construction process, which differs significantly from the traditional appending-based approaches, we redefine the SL-based training scheme as follows.

For a problem instance with $n$ nodes and its labeled solution $\boldsymbol{\pi}^*$ (e.g., an optimal tour), we first randomly select a node as the initial partial solution while designating the remaining nodes as unvisited. Guided by the original labeled solution, the model then learns to sequentially re-insert these unvisited nodes into their target positions within the evolving partial solution. The target position for an unvisited node is defined by the two visited nodes in the current partial solution that are adjacent to this unvisited node in the complete labeled solution. For example, consider the partial solution illustrated in Figure~\ref{Training example}, where $\pi_2$ represents the current node. After excluding the unvisited nodes $\pi_4$ and $\pi_5$, the two visited nodes adjacent to $\pi_2$ in the labeled solution are $\pi_1$ and $\pi_3$. Therefore, the target position is $\vec{e}^{\,*} =\vec{e}_1 = (\pi_1, \pi_3)$. The loss function is then computed as:
\begin{equation}
\label{eq: loss function}
\mathcal{L}(\boldsymbol{\theta}) =  - \log p_{\boldsymbol{\theta}}\left(\vec{e}^{\,*}  \mid  \vec{e}_{\pi_{1: l}},\boldsymbol{\pi}^*\right),
\end{equation}
where $\{ \vec{e}_{\pi_{1:l}}\}$ is the set of positions in the current incomplete solution, $\vec{e}^{\,\ast} \in \{ \vec{e}_{\pi_{1:l}}\}$ is the target position, and $p_{\boldsymbol{\theta}}\left(\vec{e}^{\,\ast} \mid  \vec{e}_{\pi_{1: l}}, S\right)$ is the probability of inserting the current node into target psition predicted by the model parameterized by $\boldsymbol{\theta}$. Subsequently, the model parameters are optimized using a gradient-based algorithm (e.g., Adam ~\citep{Adam}) to minimize the loss function. Through this training process, the model progressively learns to accurately position previously unvisited nodes into their correct positions within the solution sequence, ultimately acquiring the capability to generate high-quality solutions.

\subsection{Insertion-based Local Reconstruction}
\label{subsection:Insertion-based Local Reconstruction}

For NCO, local reconstruction techniques are widely employed during the inference phase to improve solution quality through iterative refinement of local solution segments~\citep{hottung2020neural,kim2021learning,luo2023neural,ye2024glop}. While these methods can effectively trade computational time for enhanced solution quality, they face inherent limitations when applied to appending-based constructive models. Specifically, the constraint of using sequence-based destruction strategies severely limits both the destruction space and the subsequent reconstruction action space, often trapping solutions in local optima. For example, consider a suboptimal solution wherein two nodes are neighbors in terms of the geometric distance but are distant in their relative sequence positions. Under sequence-based deconstruction, such nodes have a low probability of being in the same destruction space. Consequently, if an optimal solution requires these two nodes to be adjacent, the subsequent reconstruction step is unlikely to reconnect them, thereby causing the suboptimal solution to stall in a local optimum.

To overcome this limitation, we propose an insertion-based local reconstruction method that exploits the inherent flexibility of the insertion paradigm through a distance-based deconstruction strategy, which can enable more effective reconstruction and consequently enhance the model performance. The iterative local reconstruction process alternates between destruction and reconstruction phases. 
\textbf{(1) Destruction Phase:} The solution is destroyed based on the distance-based proximity of nodes. The process begins by randomly selecting a node from the current solution, followed by the identification and removal of its $\alpha$-nearest neighbors, which transition to unvisited nodes. The remaining solution segments automatically form a new reduced loop solution. Here, $\alpha$ represents the destruction size. For example, given a TSP10 solution $(\pi_1, \pi_2, \ldots, \pi_{10})$ where $\pi_5$ is randomly selected with its 3-nearest neighbors $\{\pi_2, \pi_7, \pi_8\}$, the removal of $\pi_5$ and $\{\pi_2, \pi_7, \pi_8\}$ yields the looped solution $(\pi_1, \pi_3, \pi_4, \pi_6, \pi_9, \pi_{10})$. 
\textbf{(2) Reconstruction Phase:} The model sequentially inserts the unvisited nodes into this partial solution. Once all nodes are re-inserted,  the newly generated solution is compared against its predecessor, and the superior solution is retained for subsequent iterations. This iterative process continues until a predefined computational budget (e.g., a maximum number of iterations) is exhausted.

\begin{table}[t]
  \centering
  \caption{Results on synthetic TSP and CVRP instances. Results marked with an asterisk (*) are from the original paper. The best results are bold. OOM: The method triggered an out-of-memory condition. The reported inference time is the total time required to solve the entire test set.}
  \resizebox{\textwidth}{!}{%
    \begin{tabular}{l|ccc|ccc|ccc}
    \toprule[0.5mm]
    \multirow{2}[4]{*}{Method} & \multicolumn{3}{c|}{TSP100} & \multicolumn{3}{c|}{TSP1K} & \multicolumn{3}{c}{TSP10K} \\
& Length & Gap   & Time  & Length & Gap   & Time  & Length & Gap   & Time \\
    \midrule
        LKH3  & 7.76325 & 0.000\% & 56m   & 23.119 & 0.000\% & 8.2h  & 71.778 & 0.000\% & 8.8h \\
    Concorde & 7.76324 & 0.000\% & 34m   & 23.118 &-0.003\% & 7.8h  & 72    & 0.310\% & 22.4h \\

    \midrule
    Att-GCN+MCTS* & 7.764 & 0.037\% & 15m   & 23.863 & 3.224\% & 13m   & 74.93 & 4.390\% & 1.8h \\
    DIFUSCO* & 7.78  & 0.240\% & -     & 23.56 & 1.900\% & 12.1m & 73.62 & 2.580\% & 47m \\
    T2T*  & 7.76  & 0.060\% & -     & 23.3  & 0.780\% & 54.7m & -     & -     & - \\
    \midrule
    NeurOpt &  7.765   &  0.023\%   &   1.8h    &   307.578    &   > 100\%   &    3h   &   -   &   -    &  - \\
    SO-mixed* & -     & -     & -     & 23.766 & 2.800\% & 55.5m & 74.299 & 3.517\% & 2.034h \\
    H-TSP & -     & -     & -     & 24.718 & 6.912\% & 34s   & 77.75 & 8.320\% & 25m \\
    GLOP  & 7.767 & 0.048\% & 1.4h  & 23.84 & 3.119\% & 2.4m  & 75.04 & 4.545\% & 25m \\
    UDC-$\textbf{x}_{50}$ ($\alpha=50$)  & 7.788 &  0.319\% & 1.38h & 23.53 &  1.782\% & 6.21m & OOM & OOM & OOM  \\
    \midrule
    POMO aug×8 & 7.774 & 0.138\% & 0.7m    & 32.5  & 40.577\% & 6.1m  & OOM   & OOM   & OOM \\
    INViT-3V aug$\times$16  & 7.92 &  2.019\% & 3.02h & 24.343 &  5.299\% & 30.43m & 76.494 &  6.570\% & 1.19h  \\
    ELG   & 7.781 & 0.229\% & 2.2m  & 25.738 & 11.328\% & 1.3m  & OOM   & OOM   & OOM \\
    BQ bs16 & 7.764 & 0.010\% & 20m   & 23.432 & 1.354\% & 21m   & OOM   & OOM   & OOM \\
    LEHD RRC1000 & 7.76337 & 0.0016\% & 2.3h  & 23.288 & 0.731\% & 5.2h  & 80.9  & 12.709\% & 3.6h \\
    \midrule
    L2C-Insert Greedy & 7.799  & 0.459\% &  0.86m & 24.218 &  4.754\% & 0.35m & 77.343   & 7.754\% & 1.06m \\
    L2C-Insert ($I$=100) & 7.764  & 0.0038\% & 47.9m  & 23.414   & 1.275\% & 4.79m & 75.603 & 5.329\% & 2.58m \\
    L2C-Insert ($I$=200) & 7.76336  & 0.0014\% & 1.62h  & 23.339 &  0.952\% & 9.52m & 74.733  & 4.117\% & 4.14m \\
    L2C-Insert ($I$=500) & 7.76328 & 0.0005\% & 4.0h  & 23.265  &  0.633\% & 22.94m & 73.812  & 2.834\% & 8.73m  \\
    L2C-Insert ($I$=1000) & \textbf{7.76326}  & \textbf{0.0002\%} &  8.0h & \textbf{23.230}   &  \textbf{0.481\%} & 46.4m & \textbf{73.270} & \textbf{2.079\%} & 16.64m \\

    \midrule
    \midrule
    \multirow{2}[4]{*}{Method} & \multicolumn{3}{c|}{CVRP100} & \multicolumn{3}{c|}{CVRP1K}& \multicolumn{3}{c}{CVRP10K} \\
          & Length & Gap   & Time  & Length & Gap   & Time  & Length & Gap   & Time \\
    \midrule
    HGS   & 15.56 & 0.000\%     & 4.5h  & 41.161 & 0.000\% & 5h & 226.177
      &    0.000\%   & 8h \\
    LKH3  & 15.65 & 0.540\% & 12h   & 42.156  & 2.416\% & 6.1h &  290.575     &   28.472\%    & 10.3h \\
    \midrule
    NeurOpt &   15.66    &    0.627\%   &   3h   &   -   &   -    &   -    &   -    &    -   & - \\
    GLOP-G (LKH3) & -     & -     & -     &  45.753  &   11.154\%   & 1.93m  &  276.252  &  22.140\% & 27.98m \\
    UDC-$\textbf{x}_{50}$ ($\alpha=50$) & 16.291 &  4.699\% &  2.69h & 43.583 &  5.884\% & 11.96m & OOM &  OOM &  OOM \\
    \midrule
    POMO aug×8 & 15.75 & 1.230\% & 0.86m  &  100.990 & >100\% & 5.83m &   OOM    &  OOM     & OOM \\
    INViT-3V aug$\times$16  & 16.538 &  6.282\% &  3.93h & 46.607 &  13.230\% & 26.97m & 275.691 &  21.892\% & 1.28h \\
    ELG   & 15.84 & 1.760\% & 3.4m  & 46.847 &  13.812\% & 3.16m & 277.685  & 22.773\%   & 11.92m \\
    BQ bs16 & 15.81 & 1.560\% & 18m   & 43.121 & 4.762\% & 15m & 275.494 & 21.805\% & 1.36h \\
    LEHD RRC1000 & 15.63 & 0.420\% & 2.0h  &   \textbf{42.408}  &  \textbf{3.028\%} &  4.15h &  248.604 &   9.916\%   &  8.1h \\
    \midrule
    L2C-Insert Greedy & 16.169  & 3.892\% & 1.4m  & 44.458   & 8.009\% & 1.17m  &    255.356  &   12.901\%    & 2.88m \\
    L2C-Insert ($I$=100) & 15.700  & 0.877\% & 1.0h  & 43.868   & 6.574\% &  12.08m&   253.462   &    12.063\%   &  5.75m \\
    L2C-Insert ($I$=200) & 15.667  & 0.664\% & 2.0h  & 43.677 & 6.113\% & 24.66m &    252.021   &   11.426\%    & 9.02m  \\
    L2C-Insert ($I$=500) & 15.640  & 0.493\% & 5.3h  &  43.416 & 5.478\% &  1.05h &   248.987    &  10.085\%     &  19.03m\\
    L2C-Insert ($I$=1000) & \textbf{15.627}  & \textbf{0.413\%} & 10.6h & 43.152   & 4.836\% & 2.05h &    \textbf{245.732 }	& \textbf{8.646\%}	& 34.72m\\
    \bottomrule[0.5mm]
    \end{tabular}%
    }
  \label{main-table1}%
\end{table}%

\begin{table}[htbp]
  \centering
  \caption{Results on synthetic TSP and CVRP instances with 50K/100K nodes. The best results are bold. 
  }
    \resizebox{0.75\textwidth}{!}{%
    \begin{tabular}{l|ccc|ccc}
    \toprule
    \multirow{2}[2]{*}{Method} & \multicolumn{3}{c|}{TSP50K} & \multicolumn{3}{c}{TSP100K} \\
          & Length & Gap   & Time  & Length & Gap   & Time \\
    \midrule
    LKH3  & 159.93 & 0.000\%     & 160h   & 225.99 & 0.000\%   & 400h \\
    \midrule
    GLOP  & 168.09 & 5.102\% & 24m   & 237.61 & 5.142\% & 1.04h \\
    INViT greedy & 171.42 & 7.184\% & 20.8h & 242.26 & 7.199\% & 80h \\
      LEHD RRC1000 &  OOM & OOM  &  OOM &  OOM & OOM  &  OOM \\
    \midrule
    L2C-Insert Greedy & 171.778 & 7.408\%  & 5.28m& 242.757 & 7.419\%  & 10.53m \\ 
    L2C-Insert ($I$=200) & 169.802 & 6.173\%  & 8.35m & 241.194 & 6.728\%  & 13.60m \\
    L2C-Insert ($I$=1000) & \textbf{166.063} & \textbf{3.835\%} & 20.83m & \textbf{237.109} & \textbf{4.920\%}  & 26.07m \\
    \midrule
    \multirow{2}[2]{*}{Method} & \multicolumn{3}{c|}{CVRP50K} & \multicolumn{3}{c}{CVRP100K} \\
          & Length & Gap   & Time  & Length & Gap   & Time \\
    \midrule
    HGS   & 1081.0  & 0.000\%     & 64h   & 2087.5 & 0.000\%     & 100.8h \\
    \midrule
    INViT-3V greedy & 1331.1 & 23.136\% & 46.4h & 2683.4 & 28.546\% & 93h \\
        BQ bs16 &  OOM & OOM  &  OOM &  OOM & OOM  &  OOM \\
    LEHD RRC1000 &  OOM & OOM  &  OOM &  OOM & OOM  &  OOM \\
    \midrule
    L2C-Insert Greedy &   1188.084    &   9.906\%    & 14.49m    & 2271.292 & 8.804\% & 29.95m \\
    L2C-Insert ($I$=200) &   1184.744    &   9.597\%    &  20.90m    & 2268.146 & 8.654\% & 37.67m \\
    L2C-Insert  ($I$=1000) & \textbf{1174.208}      & \textbf{8.622\%}      &   47.63m    & \textbf{2258.005}
 & \textbf{8.168\%} & 1.16h \\
    \bottomrule
    \end{tabular}%
    }
  \label{tab: main result 2}%
\end{table}%

\section{Experiments}
\label{Section: Experiment}

In this section, we empirically evaluate the L2C-Insert framework on both synthetic and real-world benchmark TSP and CVRP datasets of various scales. We then contact ablation studies to verify the effect of the proposed framework’s critical components.

\paragraph{Datasets}

Following the standard data generation procedure from prior work~\cite{kool2018attention}, we generate datasets for both TSP and CVRP problems. \textbf{(1) Training Datasets:} For each problem (TSP and CVRP), we create training datasets consisting of one million 100-node instances. We obtain (near-)optimal solutions for these instances using Concorde~\cite{applegate2006concorde} for TSP and HGS~\citep{HGS} for CVRP. \textbf{(2) Test Datasets:}  Consistent with established  practices~\citep{fu2021generalize,luo2023neural,hou2023generalize,luo2024self}, we generate five synthetic test datasets with instance sizes ranging from 100 to 100K nodes (100, 1K, 10K, 50K, and 100K). We denote TSP and CVRP instances with $N$ nodes as TSP$N$ and CVRP$N$ (e.g., TSP100, CVRP100). The dataset composition is as follows:
For TSP, the TSP100 and TSP1K datasets contain 10,000 and 128 instances, respectively, while larger-scale datasets (10K, 50K, 100K nodes) each have 16 instances. CVRP datasets follow a similar structure, with the exception that CVRP1K contains 100 instances, as specified in~\citep{hou2023generalize}. Following \citet{hou2023generalize}, we set vehicle capacities to 50 for CVRP100, 200 for CVRP1K, and 300 for larger-scale instances. To evaluate our method on real-world large-scale instances, we also collect 49 symmetric TSP instances and 100 CVRP instances with EUC\_2D features and $\leq$1K nodes from TSPLIB~\cite{Rein91} and CVRPLIB Set-X~\cite{uchoa2017new}, respectively.

\paragraph{Model Setting \& Training}

The embedding dimension is set to $d=128$, the hidden dimension of the feed-forward layer is set to 512, the query dimension $d_q$ is set to 16, and the head number in multi-head attention is set to 8. We set the number of attention layers in the decoder to $L=9$ following~\citep{drakulic2023bq}. Both the TSP and CVRP models are trained on one million instances of size 100. For TSP model training, we employ the Adam optimizer~\citep{Adam}. For CVRP model training, we employ the AdamW~\citep{loshchilov2018decoupled} optimizer, as we find it to be more stable. The initial learning rate is 1×10$^{\mbox{-}4}$, decaying by a factor of 0.97 after each epoch. The TSP and CVRP models undergo 50 and 15 training epochs, respectively, with a batch size of 1024. All experiments, including training, testing, and evaluation, are conducted on a single NVIDIA GeForce RTX 4090 GPU with 24 GB of memory to ensure consistent computational conditions.

\paragraph{Baselines}
We compare L2C-Insert with a comprehensive set of baseline methods including \textbf{(1) Classical Solvers:} Concorde~\citep{applegate2006concorde}, LKH3~\citep{LKH3}, and HGS~\cite{HGS}; \textbf{(2) Constructive NCO Methods:} POMO~\citep{kwon2020pomo}; BQ~\citep{drakulic2023bq}, LEHD~\citep{luo2023neural}, INViT~\citep{fang2024invit}, and ELG~\citep{gao2024towards}. (3) \textbf{Heatmap-based NCO Methods:} Att-GCN+MCTS~\citep{fu2021generalize}, DIFUSCO~\citep{sun2023difusco}, and T2T~\citep{li2023from}; (4) \textbf{Two-stage NCO Methods:} H-TSP~\citep{pan2023h-tsp}, SO-mixed~\citep{cheng2023select}, GLOP~\citep{ye2024glop}, and UDC~\citep{ye2024glop}; (5) \textbf{Improvement-based NCO Methods:} NeurOpt~\citep{ma2023neuopt}. In line with the established practice in the NCO literature~\citep{kool2018attention,kwon2020pomo}, the classical solvers are primarily included to provide a strong, near-optimal baseline for solution quality, rather than to serve as direct competitors in terms of runtime. We refer to Appendix~\ref{Implementation details of Baselines} for implementation details of these baselines.

\paragraph{Inference \& Metrics}

For inference, the L2C-Insert model first generates an initial solution through greedy rollout, then progressively refines solution quality via insertion-based local reconstruction. The number of iterations is denoted as ``$I$=''. Following~\citep{drakulic2023bq}, we accelerate inference by processing only the $k$-nearest unvisited nodes and positions of the current node, with $k$ set to 100 for TSP and 200 for CVRP. To further accelerate this process, we limit the local reconstruction destruction size $\alpha$ to 300 across all instances. Following~\citep{ye2024glop}, to enhance the consistency of inputs for our model on large-scale instances with more than 1K nodes, we employ min-max feature scaling, which adjusts the x-axis and y-axis coordinate values to the interval [0, 1]. A hyperparameter study and discussion on these settings are provided in Appendix~\ref{appd: hyperparameter study}. For performance evaluation, we measure: 1) average tour length (Length), 2) performance gap relative to baseline solvers (Gap), and 3) total inference time (Time). Note that reported inference times for classical CPU-based solvers are not directly comparable to GPU-executed learning-based methods.

\begin{table}[t]
\centering
\caption{Comparison result for appending- and insertion-based constructive heuristics on TSPLIB and CVRPLIB instances.
}
\label{main table-LIB}
\resizebox{0.85\textwidth}{!}{
        \begin{tabular}{l|ccc|ccc}
    \toprule
    \multicolumn{1}{l|}{\multirow{2}[1]{*}{Method}} & \multicolumn{3}{c|}{TSPLIB} & \multicolumn{3}{c}{CVRPLIB} \\
& N<=200 & 200<N<=1K & All  & N<=200 & 200<N<=1K & All \\
    \midrule
    NeurOpt  & 9.456\% & 26.345\% & 16.350\%  & 24.878\% & 25.301\% & 25.190\%  \\
    UDC-$\textbf{x}_{250}$ ($\alpha=50$) & 0.533\% & 6.578\% & 3.072\%  & 9.635\%  & 8.864\% & 9.033\%  \\
       GLOP  & 0.561\%  & 2.012\% &  1.153\% & - & - & -  \\
    INViT-3V aug$\times$16  & 2.238\% & 5.654\% &  3.673\% &  8.231\% & 11.504\% & 10.784\%  \\
   ELG  & 1.180\%  & 5.910\% & 3.111\%  &  4.510\%   & 7.752\% & 7.039\%  \\
   POMO aug$\times$8 & 2.022\%  & 21.447\% & 9.951\%  & 6.869\% & 46.984\% & 38.159\% \\
   BQ bs16  & 1.289\%  & 1.946\% & 1.557\%  &    7.661\% & 8.195\% & 8.078\% \\
    LEHD RRC1000  &  0.215\% &  0.620\%  & 0.380\%   & 2.022\% & 3.932\% & 3.512\% \\
    \midrule
    L2C-Insert ($I$=1000)   & \textbf{0.197\%}  & \textbf{0.468\%} & \textbf{0.308\%}  &  \textbf{1.604\%} & \textbf{3.651\% } & \textbf{3.201\%} \\
    \midrule
    \end{tabular}%
}
\end{table}

\subsection{Comparative Results}

Tables~\ref{main-table1} and~\ref{tab: main result 2} present the comparative results on uniformly distributed TSP and CVRP instances scaling from 100 to 100K nodes. These results show that L2C-Insert consistently achieves outstanding performance across these scales with a reasonable runtime. On TSP100, L2C-Insert ($I$=200) reduces the gap of the advanced NCO baseline LEHD+RRC1000 by 12.5\% with a speedup of $1.4\times$. On TSP1K, L2C-Insert ($I$=1000) achieves a significant reduction of 34.2\% while with a speedup of 6.7$\times$ compared to LEHD+RRC1000. On TSP10K, L2C-Insert ($I$=1000) outperforms the strong heatmap-based solver DIFUSCO and two-stage solver GLOP with 19.4\% and 54.3\% gap reduction, while with 2.8$\times$ and 1.5$\times$ speedup, respectively. On TSP50K and TSP100K, L2C-Insert ($I$=200) outperforms the two-stage NCO method GLOP with a gap reduction of 24.8\% and 4.3\%, respectively. On CVRP100, L2C-Insert ($I$=1000) outperforms LEHD RRC1000 by 1.7\%, and on CVRP1K it achieves a competitive gap while with a speedup of 2$\times$. On CVRP10K, L2C-Insert ($I$=1000) outperforms LEHD RRC1000 with a speed-up of 14$\times$. On CVRP50K and CVRP100K, L2C-Insert using greedy rollout achieves significant gap reductions of 62.7\% and 71.4\%, respectively, compared to INViT. Overall, our method shows good scalability and can achieve outstanding performance across small-scale to very large-scale problem instances with up to 100K nodes. The experimental results on real-world TSPLIB and CVRPLIB instances are provided in Table~\ref{main table-LIB}. These results show that L2C-Insert consistently outperforms representative neural solvers on all instance groups, demonstrating robust generalization ability.

\subsection{Ablation Study}
\label{Ablation Study}

\paragraph{Learning to Append vs. Learning to Insert}

We conduct a comparative evaluation between two constructive NCO approaches: learning to append (L2C-Append) and learning to insert (L2C-Insert). While L2C-Append shares a similar model architecture with L2C-Insert, it utilizes an appending construction framework akin to LEHD~\citep{luo2023neural}. Both models are trained on TSP100 under identical budgets and subsequently evaluated on TSP100, TSP1K, and TSP10K datasets. For inference, we apply local reconstruction with 200 and 1000 iterations using the aforementioned settings. As detailed in Table~\ref{abla: L2C-Append vs. L2C-Insert}, L2C-Insert, with only 200 reconstruction iterations, outperforms L2C-Append even when the latter uses 1000 iterations of local reconstruction while also consuming less runtime. Remarkably, L2C-Insert ($I$=1000) outperforms L2C-Append ($I$=1000) by achieving substantial gap reductions of at least 50.6\% on these datasets. These findings highlight that the L2C-Insert framework's greater flexibility in solution construction directly contributes to its superior performance.

\begin{table}[htbp]
  \centering
  \caption{Appending-based Construction vs. Insertion-based Construction. The best gap is indicated in bold, and the second-best is underlined.}
  \resizebox{0.8\textwidth}{!}{
    \begin{tabular}{cc|cc|cc|cc}
    \toprule
    \multicolumn{2}{c|}{\multirow{2}[2]{*}{Method}} & \multicolumn{2}{c|}{TSP100} & \multicolumn{2}{c|}{TSP1000} & \multicolumn{2}{c}{TSP10000} \\
    \multicolumn{2}{c|}{} & Gap   & Time  & Gap   & Time  & Gap   & Time \\
    \midrule
    L2C-Append      & ($I$=200) & 0.00489\% & 28.03m & 1.159\% & 3.49m & 22.986\% & 4.11m \\
          & ($I$=1000) & 0.00144\% & 2.28h & 0.974\% & 16.88m & 12.881\% & 16.30m \\
    \midrule
        L2C-Insert   & ($I$=200) & \underline{0.00142\%} & 1.62h & \underline{0.952\%} & 9.52m & \underline{4.117\%} & 4.14m \\
          & ($I$=1000) & \textbf{0.00017\%} & 8h& \textbf{0.481\%} & 46.4m & \textbf{2.079\%} & 16.64m \\
    \bottomrule
    \end{tabular}%
    }
  \label{abla: L2C-Append vs. L2C-Insert}%
\end{table}%

\begin{wraptable}{r}{0.4\textwidth} 
\centering
\caption{Effect of deconstruction strategy.}
\vspace{7pt}
\label{Effect of destroy strategy}
\resizebox{0.4\columnwidth}{!}{
  \begin{tabular}{l | c c }
    \toprule[0.5mm]
    & TSP1000 & TSP10000    \\ 
  Method  &  Gap   &  Gap   \\ 
    \midrule
Sequence-based    &    1.182\%   &    6.980\%       \\
Distance-based  &    \textbf{0.481\%}    &   \textbf{2.079\%}     \\
    \bottomrule[0.5mm]
    \end{tabular}%
}
\end{wraptable}

\paragraph{Effect of Destruction Strategy}

We ablate the effect of distance-based strategies on local reconstruction. Specifically, on TSP1K and TSP10K, we test the L2C-Insert model using local reconstruction ($I$=1000) with the distance-based and sequence-based destruction separately. As demonstrated in Table~\ref{Effect of destroy strategy}, the distance-based destruction strategy achieves substantially better performance, indicating its greater effectiveness in helping local reconstruction escape local optima.

\paragraph{Additional Analysis \& Discussion}

We further investigate the impact of: 1) different distinct distance metrics (Euclidean vs. Polar Angle); 2) the use of different node selection strategies (Random vs. Nearest); 3) the incorporation of Positional Encodings~\citep{vaswani2017attention} into embeddings of positions within the partial solution; 4) the exclusion of unvisited nodes as decoder input; and 5) the use of different backbone models for L2C-Insert framework. Detailed results and the corresponding discussion are provided in Appendix~\ref{Additional Ablation Studies}.

\section{Conclusion, Limitation, and Future Work}
\label{Conclusion}

\paragraph{Conclusion} In this work, we have proposed L2C-Insert, a novel learning to insert framework to explore the potential of the insertion paradigm for constructive NCO. By strategically inserting unvisited nodes at any valid position within the partial solution, L2C-Insert can flexibly construct high-quality solutions for routing problems. Extensive comparisons with advanced neural solvers demonstrate L2C-Insert's outstanding performance on TSP and CVRP. 

\paragraph{Limitation and Future Work} First, although L2C-Insert has demonstrated outstanding performance, its performance could be further boosted through the development of more effective model architectures. These improved architectures would better capture the intricate relationships between current nodes, unvisited nodes, and positions within partial solutions. Second, designing a hybrid NCO solver that combines both appending and inserting moves is a highly promising research direction. Such a model has the potential to be more flexible and powerful than a model restricted to a single type of move. In addition, a limitation may be the greedy, non-learned nearest-neighbor heuristic used for node selection. A promising research direction is to develop a learned policy that works synergistically with the insertion model, potentially through joint or two-stage training. Finally, extending the L2C-Insert framework to address a broader range of combinatorial optimization problems is another promising research direction for future work.

\begin{ack}

This work was supported in part by the National Natural Science Foundation of China (Grant 62476118, Grant 12426305), the Natural Science Foundation of Guangdong Province (Grant 2024A1515011759), the Guangdong Science and Technology Program (Grant 2024B1212010002), the Center for Computational Science and Engineering at Southern University of Science and Technology, the National Key Research and Development Program of China (Grant 2022YFA1004201), and the Research Grants Council of the Hong Kong Special Administrative Region, China (GRF Project No. CityU 11212524).
\end{ack}

{\small

\bibliographystyle{unsrtnat}
\bibliography{reference}        
}

\clearpage

\section*{NeurIPS Paper Checklist}

\begin{enumerate}

\item {\bf Claims}
    \item[] Question: Do the main claims made in the abstract and introduction accurately reflect the paper's contributions and scope?
    \item[] Answer: \answerYes{} 
    \item[] Justification: Our abstract clearly reflects the scope (Constructive Neural Combinatorial Optimization), and our introduction clearly summarizes our contributions.
    \item[] Guidelines:
    \begin{itemize}
        \item The answer NA means that the abstract and introduction do not include the claims made in the paper.
        \item The abstract and/or introduction should clearly state the claims made, including the contributions made in the paper and important assumptions and limitations. A No or NA answer to this question will not be perceived well by the reviewers. 
        \item The claims made should match theoretical and experimental results, and reflect how much the results can be expected to generalize to other settings. 
        \item It is fine to include aspirational goals as motivation as long as it is clear that these goals are not attained by the paper. 
    \end{itemize}

\item {\bf Limitations}
    \item[] Question: Does the paper discuss the limitations of the work performed by the authors?
    \item[] Answer: \answerYes{} 
    \item[] Justification: The paper discusses the limitations of the work in Section~\ref{Conclusion}. 
    \item[] Guidelines:
    \begin{itemize}
        \item The answer NA means that the paper has no limitation while the answer No means that the paper has limitations, but those are not discussed in the paper. 
        \item The authors are encouraged to create a separate "Limitations" section in their paper.
        \item The paper should point out any strong assumptions and how robust the results are to violations of these assumptions (e.g., independence assumptions, noiseless settings, model well-specification, asymptotic approximations only holding locally). The authors should reflect on how these assumptions might be violated in practice and what the implications would be.
        \item The authors should reflect on the scope of the claims made, e.g., if the approach was only tested on a few datasets or with a few runs. In general, empirical results often depend on implicit assumptions, which should be articulated.
        \item The authors should reflect on the factors that influence the performance of the approach. For example, a facial recognition algorithm may perform poorly when image resolution is low or images are taken in low lighting. Or a speech-to-text system might not be used reliably to provide closed captions for online lectures because it fails to handle technical jargon.
        \item The authors should discuss the computational efficiency of the proposed algorithms and how they scale with dataset size.
        \item If applicable, the authors should discuss possible limitations of their approach to address problems of privacy and fairness.
        \item While the authors might fear that complete honesty about limitations might be used by reviewers as grounds for rejection, a worse outcome might be that reviewers discover limitations that aren't acknowledged in the paper. The authors should use their best judgment and recognize that individual actions in favor of transparency play an important role in developing norms that preserve the integrity of the community. Reviewers will be specifically instructed to not penalize honesty concerning limitations.
    \end{itemize}

\item {\bf Theory assumptions and proofs}
    \item[] Question: For each theoretical result, does the paper provide the full set of assumptions and a complete (and correct) proof?
    \item[] Answer: \answerNA{} 
    \item[] Justification: The claim of this paper is supported by experimental results. 
    \item[] Guidelines:
    \begin{itemize}
        \item The answer NA means that the paper does not include theoretical results. 
        \item All the theorems, formulas, and proofs in the paper should be numbered and cross-referenced.
        \item All assumptions should be clearly stated or referenced in the statement of any theorems.
        \item The proofs can either appear in the main paper or the supplemental material, but if they appear in the supplemental material, the authors are encouraged to provide a short proof sketch to provide intuition. 
        \item Inversely, any informal proof provided in the core of the paper should be complemented by formal proofs provided in appendix or supplemental material.
        \item Theorems and Lemmas that the proof relies upon should be properly referenced. 
    \end{itemize}

    \item {\bf Experimental result reproducibility}
    \item[] Question: Does the paper fully disclose all the information needed to reproduce the main experimental results of the paper to the extent that it affects the main claims and/or conclusions of the paper (regardless of whether the code and data are provided or not)?
    \item[] Answer: \answerYes{} 
    \item[] Justification: 
 We provide the detailed experiment setting in Section~\ref{Section: Experiment}, hyperparameter settings in Appendix~\ref{appd: hyperparameter study}, and implementation in Appendix~\ref{Implementation Details of L2C-Insert}, which is fully sufficient for reproducibility.
    \item[] Guidelines:
    \begin{itemize}
        \item The answer NA means that the paper does not include experiments.
        \item If the paper includes experiments, a No answer to this question will not be perceived well by the reviewers: Making the paper reproducible is important, regardless of whether the code and data are provided or not.
        \item If the contribution is a dataset and/or model, the authors should describe the steps taken to make their results reproducible or verifiable. 
        \item Depending on the contribution, reproducibility can be accomplished in various ways. For example, if the contribution is a novel architecture, describing the architecture fully might suffice, or if the contribution is a specific model and empirical evaluation, it may be necessary to either make it possible for others to replicate the model with the same dataset, or provide access to the model. In general. releasing code and data is often one good way to accomplish this, but reproducibility can also be provided via detailed instructions for how to replicate the results, access to a hosted model (e.g., in the case of a large language model), releasing of a model checkpoint, or other means that are appropriate to the research performed.
        \item While NeurIPS does not require releasing code, the conference does require all submissions to provide some reasonable avenue for reproducibility, which may depend on the nature of the contribution. For example
        \begin{enumerate}
            \item If the contribution is primarily a new algorithm, the paper should make it clear how to reproduce that algorithm.
            \item If the contribution is primarily a new model architecture, the paper should describe the architecture clearly and fully.
            \item If the contribution is a new model (e.g., a large language model), then there should either be a way to access this model for reproducing the results or a way to reproduce the model (e.g., with an open-source dataset or instructions for how to construct the dataset).
            \item We recognize that reproducibility may be tricky in some cases, in which case authors are welcome to describe the particular way they provide for reproducibility. In the case of closed-source models, it may be that access to the model is limited in some way (e.g., to registered users), but it should be possible for other researchers to have some path to reproducing or verifying the results.
        \end{enumerate}
    \end{itemize}

\item {\bf Open access to data and code}
    \item[] Question: Does the paper provide open access to the data and code, with sufficient instructions to faithfully reproduce the main experimental results, as described in supplemental material?
    \item[] Answer: \answerYes{}
    \item[] Justification: See the abstract for the link to the code and trained models.
    \item[] Guidelines:
    \begin{itemize}
        \item The answer NA means that paper does not include experiments requiring code.
        \item Please see the NeurIPS code and data submission guidelines (\url{https://nips.cc/public/guides/CodeSubmissionPolicy}) for more details.
        \item While we encourage the release of code and data, we understand that this might not be possible, so “No” is an acceptable answer. Papers cannot be rejected simply for not including code, unless this is central to the contribution (e.g., for a new open-source benchmark).
        \item The instructions should contain the exact command and environment needed to run to reproduce the results. See the NeurIPS code and data submission guidelines (\url{https://nips.cc/public/guides/CodeSubmissionPolicy}) for more details.
        \item The authors should provide instructions on data access and preparation, including how to access the raw data, preprocessed data, intermediate data, and generated data, etc.
        \item The authors should provide scripts to reproduce all experimental results for the new proposed method and baselines. If only a subset of experiments are reproducible, they should state which ones are omitted from the script and why.
        \item At submission time, to preserve anonymity, the authors should release anonymized versions (if applicable).
        \item Providing as much information as possible in supplemental material (appended to the paper) is recommended, but including URLs to data and code is permitted.
    \end{itemize}

\item {\bf Experimental setting/details}
    \item[] Question: Does the paper specify all the training and test details (e.g., data splits, hyperparameters, how they were chosen, type of optimizer, etc.) necessary to understand the results?
    \item[] Answer: \answerYes{} 
    \item[] Justification: We provide the detailed experiment setting in Section~\ref{Section: Experiment}, hyperparameter settings in Appendix~\ref{appd: hyperparameter study}, and implementation in Appendix~\ref{Implementation Details of L2C-Insert}.
    \item[] Guidelines:
    \begin{itemize}
        \item The answer NA means that the paper does not include experiments.
        \item The experimental setting should be presented in the core of the paper to a level of detail that is necessary to appreciate the results and make sense of them.
        \item The full details can be provided either with the code, in appendix, or as supplemental material.
    \end{itemize}

\item {\bf Experiment statistical significance}
    \item[] Question: Does the paper report error bars suitably and correctly defined or other appropriate information about the statistical significance of the experiments?
    \item[] Answer: \answerNo{} 
    \item[] Justification: In the field of NCO, the performance of methods is typically not evaluated by error bars or statistical significance. Instead, NCO methods conventionally keep the same seed (seed=123 in our work) during training/inference, and use the objective value (Obj.), optimality gap (Gap), and inference time (Time) as metrics to evaluate method performance. A detailed description of these metrics is provided in Section~\ref{Section: Experiment}.
    \item[] Guidelines:
    \begin{itemize}
        \item The answer NA means that the paper does not include experiments.
        \item The authors should answer "Yes" if the results are accompanied by error bars, confidence intervals, or statistical significance tests, at least for the experiments that support the main claims of the paper.
        \item The factors of variability that the error bars are capturing should be clearly stated (for example, train/test split, initialization, random drawing of some parameter, or overall run with given experimental conditions).
        \item The method for calculating the error bars should be explained (closed form formula, call to a library function, bootstrap, etc.)
        \item The assumptions made should be given (e.g., Normally distributed errors).
        \item It should be clear whether the error bar is the standard deviation or the standard error of the mean.
        \item It is OK to report 1-sigma error bars, but one should state it. The authors should preferably report a 2-sigma error bar than state that they have a 96\% CI, if the hypothesis of Normality of errors is not verified.
        \item For asymmetric distributions, the authors should be careful not to show in tables or figures symmetric error bars that would yield results that are out of range (e.g. negative error rates).
        \item If error bars are reported in tables or plots, The authors should explain in the text how they were calculated and reference the corresponding figures or tables in the text.
    \end{itemize}

\item {\bf Experiments compute resources}
    \item[] Question: For each experiment, does the paper provide sufficient information on the computer resources (type of compute workers, memory, time of execution) needed to reproduce the experiments?
    \item[] Answer: Yes 
    \item[] Justification: We describe the compute resources in Section~\ref{Section: Experiment}.
    \item[] Guidelines:
    \begin{itemize}
        \item The answer NA means that the paper does not include experiments.
        \item The paper should indicate the type of compute workers CPU or GPU, internal cluster, or cloud provider, including relevant memory and storage.
        \item The paper should provide the amount of compute required for each of the individual experimental runs as well as estimate the total compute. 
        \item The paper should disclose whether the full research project required more compute than the experiments reported in the paper (e.g., preliminary or failed experiments that didn't make it into the paper). 
    \end{itemize}
    
\item {\bf Code of ethics}
    \item[] Question: Does the research conducted in the paper conform, in every respect, with the NeurIPS Code of Ethics \url{https://neurips.cc/public/EthicsGuidelines}?
    \item[] Answer: \answerYes{}  
    \item[] Justification: Our research conducted in the paper conforms with the NeurIPS Code of Ethics.
    \item[] Guidelines:
    \begin{itemize}
        \item The answer NA means that the authors have not reviewed the NeurIPS Code of Ethics.
        \item If the authors answer No, they should explain the special circumstances that require a deviation from the Code of Ethics.
        \item The authors should make sure to preserve anonymity (e.g., if there is a special consideration due to laws or regulations in their jurisdiction).
    \end{itemize}

\item {\bf Broader impacts}
    \item[] Question: Does the paper discuss both potential positive societal impacts and negative societal impacts of the work performed?
    \item[] Answer: \answerYes{} 
    \item[] Justification:  As a learning-based method for solving VRPs, the proposed method does not have any specific potential negative societal impact. Detailed discussion can be found in Appendix~\ref{Broader Impacts}.
discussion.
    \item[] Guidelines:
    \begin{itemize}
        \item The answer NA means that there is no societal impact of the work performed.
        \item If the authors answer NA or No, they should explain why their work has no societal impact or why the paper does not address societal impact.
        \item Examples of negative societal impacts include potential malicious or unintended uses (e.g., disinformation, generating fake profiles, surveillance), fairness considerations (e.g., deployment of technologies that could make decisions that unfairly impact specific groups), privacy considerations, and security considerations.
        \item The conference expects that many papers will be foundational research and not tied to particular applications, let alone deployments. However, if there is a direct path to any negative applications, the authors should point it out. For example, it is legitimate to point out that an improvement in the quality of generative models could be used to generate deepfakes for disinformation. On the other hand, it is not needed to point out that a generic algorithm for optimizing neural networks could enable people to train models that generate Deepfakes faster.
        \item The authors should consider possible harms that could arise when the technology is being used as intended and functioning correctly, harms that could arise when the technology is being used as intended but gives incorrect results, and harms following from (intentional or unintentional) misuse of the technology.
        \item If there are negative societal impacts, the authors could also discuss possible mitigation strategies (e.g., gated release of models, providing defenses in addition to attacks, mechanisms for monitoring misuse, mechanisms to monitor how a system learns from feedback over time, improving the efficiency and accessibility of ML).
    \end{itemize}
    
\item {\bf Safeguards}
    \item[] Question: Does the paper describe safeguards that have been put in place for responsible release of data or models that have a high risk for misuse (e.g., pretrained language models, image generators, or scraped datasets)?
    \item[] Answer: \answerNA{} 
    \item[] Justification: This paper poses no such risks.
    \item[] Guidelines:
    \begin{itemize}
        \item The answer NA means that the paper poses no such risks.
        \item Released models that have a high risk for misuse or dual-use should be released with necessary safeguards to allow for controlled use of the model, for example by requiring that users adhere to usage guidelines or restrictions to access the model or implementing safety filters. 
        \item Datasets that have been scraped from the Internet could pose safety risks. The authors should describe how they avoided releasing unsafe images.
        \item We recognize that providing effective safeguards is challenging, and many papers do not require this, but we encourage authors to take this into account and make a best faith effort.
    \end{itemize}

\item {\bf Licenses for existing assets}
    \item[] Question: Are the creators or original owners of assets (e.g., code, data, models), used in the paper, properly credited and are the license and terms of use explicitly mentioned and properly respected?
    \item[] Answer: \answerYes{} 
    \item[] Justification: We have properly cited the assets (e.g., code, data, models) in section \ref{Licenses}.
    \item[] Guidelines:
    \begin{itemize}
        \item The answer NA means that the paper does not use existing assets.
        \item The authors should cite the original paper that produced the code package or dataset.
        \item The authors should state which version of the asset is used and, if possible, include a URL.
        \item The name of the license (e.g., CC-BY 4.0) should be included for each asset.
        \item For scraped data from a particular source (e.g., website), the copyright and terms of service of that source should be provided.
        \item If assets are released, the license, copyright information, and terms of use in the package should be provided. For popular datasets, \url{paperswithcode.com/datasets} has curated licenses for some datasets. Their licensing guide can help determine the license of a dataset.
        \item For existing datasets that are re-packaged, both the original license and the license of the derived asset (if it has changed) should be provided.
        \item If this information is not available online, the authors are encouraged to reach out to the asset's creators.
    \end{itemize}

\item {\bf New assets}
    \item[] Question: Are new assets introduced in the paper well documented and is the documentation provided alongside the assets?
    \item[] Answer: \answerNA{} 
    \item[] Justification: The datasets utilized in this paper are either sourced from prior research or generated using methodologies established in earlier studies. We have appropriately cited them.
    \item[] Guidelines:
    \begin{itemize}
        \item The answer NA means that the paper does not release new assets.
        \item Researchers should communicate the details of the dataset/code/model as part of their submissions via structured templates. This includes details about training, license, limitations, etc. 
        \item The paper should discuss whether and how consent was obtained from people whose asset is used.
        \item At submission time, remember to anonymize your assets (if applicable). You can either create an anonymized URL or include an anonymized zip file.
    \end{itemize}

\item {\bf Crowdsourcing and research with human subjects}
    \item[] Question: For crowdsourcing experiments and research with human subjects, does the paper include the full text of instructions given to participants and screenshots, if applicable, as well as details about compensation (if any)? 
    \item[] Answer: \answerNA{} 
    \item[] Justification: This paper studies the neural method for solving TSP and CVRP.
    \item[] Guidelines:
    \begin{itemize}
        \item The answer NA means that the paper does not involve crowdsourcing nor research with human subjects.
        \item Including this information in the supplemental material is fine, but if the main contribution of the paper involves human subjects, then as much detail as possible should be included in the main paper. 
        \item According to the NeurIPS Code of Ethics, workers involved in data collection, curation, or other labor should be paid at least the minimum wage in the country of the data collector. 
    \end{itemize}

\item {\bf Institutional review board (IRB) approvals or equivalent for research with human subjects}
    \item[] Question: Does the paper describe potential risks incurred by study participants, whether such risks were disclosed to the subjects, and whether Institutional Review Board (IRB) approvals (or an equivalent approval/review based on the requirements of your country or institution) were obtained?
    \item[] Answer:  \answerNA{} 
    \item[] Justification: This paper studies the neural method for solving TSP and CVRP, which does not involve crowdsourcing nor research with human subjects.
    \item[] Guidelines:
    \begin{itemize}
        \item The answer NA means that the paper does not involve crowdsourcing nor research with human subjects.
        \item Depending on the country in which research is conducted, IRB approval (or equivalent) may be required for any human subjects research. If you obtained IRB approval, you should clearly state this in the paper. 
        \item We recognize that the procedures for this may vary significantly between institutions and locations, and we expect authors to adhere to the NeurIPS Code of Ethics and the guidelines for their institution. 
        \item For initial submissions, do not include any information that would break anonymity (if applicable), such as the institution conducting the review.
    \end{itemize}

\item {\bf Declaration of LLM usage}
    \item[] Question: Does the paper describe the usage of LLMs if it is an important, original, or non-standard component of the core methods in this research? Note that if the LLM is used only for writing, editing, or formatting purposes and does not impact the core methodology, scientific rigorousness, or originality of the research, declaration is not required.
    \item[] Answer: \answerNA{} 
    \item[] Justification: \answerNA{} 
    \item[] Guidelines:
    \begin{itemize}
        \item The answer NA means that the core method development in this research does not involve LLMs as any important, original, or non-standard components.
        \item Please refer to our LLM policy (\url{https://neurips.cc/Conferences/2025/LLM}) for what should or should not be described.
    \end{itemize}

\end{enumerate}

\newpage

\appendix

\clearpage

\newpage

\section{Related Work}
\label{Appendix: Related Work}

\subsection{Constructive NCO Methods}

Constructive NCO methods typically arrange the input nodes into high-quality solutions autoregressively. This paradigm is inspired by the sequence-to-sequence framework~\citep{sutskever2014sequence} in natural language processing (NLP), where a model encodes an input sequence into an embedding and then autoregressively generates an output sequence based on this representation. 
\citet{vinyals2015pointer} pioneer this adaptation through their Pointer Networks (Ptr-Nets), which establishes the first sequence-to-sequence framework specifically for neural combinatorial optimization. In this paradigm, the neural model learns to predict the probability of appending the unvisited nodes to the partial solution until all nodes are visited. We call this paradigm Learning to Append (L2C-Append).
Subsequent research has further advanced L2C-Append NCO methods, with improvements focusing on model training techniques~\citep{bello2016neural}, extension to solve other COPs~\citet{nazari2018reinforcement}, and architectural enhancements~\citep{kool2018attention}.
Among them, the Attention Model \citep{kool2018attention} introduces the Transformer-based model structure~\cite{vaswani2017attention}, which achieves outstanding performances over various VRPs with up to 100 nodes. Various AM-based improved variants are proposed to narrow the performance gap to classical heuristic solvers~\citep{khalil2017learning,kwon2020pomo,xin2020step,xin2021multi,hottung2021efficient,kim2022sym,choo2022simulation,bi2022learning,jin2023pointerformer,luo2023neural,drakulic2023bq,xiao2023distilling,kim2021learning,bi2024learning,ye2024glop,sun2024learning,zheng2024dpn,huang2025rethinking,luo2025boosting,drakulic2025goal,zhou2025l2rlearningreducesearch}. 
For example, PolyNet~\citep{hottung2025polynet} enhances exploration by learning a diverse set of complementary solution strategies in a single model. By conditioning the solution generation process on a simple bit vector, PolyNet can generate diverse solutions without relying on handcrafted rules or forced initial moves. Nonetheless, L2C-Append NCO methods are fundamentally limited by their restriction to appending new nodes, which can lead to suboptimal solutions due to early greedy choices. In contrast, the insertion-based paradigm offers greater flexibility for VRPs by sequentially inserting unvisited nodes into any valid position~\citep{rosenkrantz1977analysis,liu2023heuristics}, potentially yielding higher-quality solutions. Despite these benefits, the insertion-based constructive approach has remained unexplored in NCO for VRPs in the last decade. In this work, we introduce L2C-Insert, a novel learning to insert framework, to investigate the potential of this paradigm for constructive NCO.

\subsection{Non-constructive NCO Methods}
This series of methods can be broadly categorized into three main groups: \textbf{(1) Heatmap-based methods}: 
These methods learn graph neural networks to predict a probability distribution, or heatmap, indicating the likelihood of each edge being part of the optimal solution for a given problem instance. Once the heatmap is obtained, it is used to guide subsequent decoding techniques such as greedy search, beam search~\citep{joshi2019efficient}, Monte Carlo Tree Search (MCTS)~\citep{fu2021generalize,qiu2022dimes,xia2024position}, 2-opt~\citep{sun2023difusco,li2023from}, or gradient search~\citep{li2024fast} to generate the solution. 
While this approach can successfully solve TSP instances of various sizes, its application to other routing problems like CVRP is generally more challenging.
\textbf{(2) Improvement-based methods}: 
These approaches iteratively refine solution quality. They employ neural models to predict regions within current solutions where heuristic operators, such as k-opt~\citep{ma2023neuopt} and swap~\citep{chen2019learning}, can be applied to improve the solution quality~\citep{d2020learning,ma2021learning,wu2021learning,li2021learning,ma2024learning,hottung2020neural}. 
(3) \textbf{Two-stage Method}: 
Two-stage methods: These methods address the problem through an iterative divide-and-conquer process. In each iteration, the problem is first decomposed into multiple subproblems using rule-based techniques or neural models. Subsequently, neural solvers (e.g., AM~\citep{kool2018attention}) or classical solvers (e.g., LKH3~\citep{LKH3} or HGS~\citep{HGS}) are used to solve these subproblems. The resulting partial solutions are then merged to form a complete solution~\citep{zong2022rbg,li2021learning,hou2023generalize,zheng2024udc,ye2024glop}.
Beyond these primary categories, other non-constructive NCO approaches include learning to augment classical solvers with neural components~\citep{xin2021neurolkh} or solving VRPs at a route level~\citep{delarue2020reinforcement}. A different approach is Neural Deconstruction Search (NDS)~\citep{hottung2025neural}. NDS uses a learned neural policy to iteratively deconstruct a solution by selecting customers to remove. These customers are then reinserted one by one using a rule-based greedy heuristic algorithm. In this paper, we develop a constructive neural model that learns the insertion-based heuristic, building solutions autoregressively.

\clearpage
\newpage

\section{Implementation Details of L2C-Insert on CVRP}
\label{Implementation Details of L2C-Insert}

\label{Implemetation details on CVRP}
\paragraph{Problem Formulation}

A CVRP instance $S$ can be represented as a graph with one depot node and $n$ customer nodes, and a vehicle with a fixed capacity $C$. Each node $i\in \{0,1,\ldots,n\}$ is associated with a feature vector $\boldsymbol{s}_i \in \mathbb{R}^{3}$, representing its two-dimensional coordinates and demand. Node 0 is the depot, and its demand is zero.

For any given CVRP instance, the solution is a set of routes for the vehicle to serve all customers, such that: (1) Each customer is visited exactly once. (2) Each route must start from and end with the depot. (3) The total demand on each route does not exceed the vehicle's capacity. The objective of solving CVRP is to determine the optimal solution such that the total cost of the routes, typically the total distance traveled, is minimized.

\paragraph{Solution Construction Process}

Initially, all customer nodes are unvisited. In the first step, the depot node and its nearest unvisited node initialize the incomplete solution. We call the node selected in the last step as \textbf{last node} and the node selected in the current step as \textbf{current node}. In each step except the first step, only the unvisited customer nodes can be selected as the current node. In each following step, the model predicts both the probability of 1) inserting the current node for each available position in the incomplete solution and 2) inserting it into the middle of the two copies of the depot nodes to initiate a new route.

\paragraph{Remaining Capacity} During solution construction, the remaining capacity associated with each route is calculated by using the vehicle's capacity minus the total demand of this route so far.

\subsection{Encoder}
The encoder transforms each node feature $\boldsymbol{s}_i\in \mathbb{R}^{3}$ to node embeddings $\mathbf{h}_i\in \mathbb{R}^{d}$ by a linear projection:
\begin{equation}
    \mathbf{h}_i^0=\boldsymbol{s}_i {W}^{(0)}+\mathbf{b}^{(0)} \quad \forall i =0,1,\ldots,n,
\end{equation}
where ${W}^{(0)}\in \mathbb{R}^{3\times d}, \mathbf{b}^{(0)} \in \mathbb{R}^{d}$ are learnable parameters. These initial embeddings are subsequently updated through the attention layer~\citep{kwon2020pomo,luo2023neural} to produce the refined embeddings $\mathbf{h}_i\in \mathbb{R}^{d}$, which is the output of the encoder.

\subsection{Decoder}

 At construction step $t$, the decoder receives the embeddings of all nodes, denoted as $\mathbf{h}_i\in \mathbb{R}^{d}, i=1,\dots,n$, the current partial solution $(0,\pi_1, \pi_2, \ldots,0,\cdots, \pi_{t-1},0)$ (Node 0 denotes the depot), and the set of unvisited nodes $\{u_j, u_j \notin \{\pi_{1:t-1}\} \}$. To compute the probability of insertion at the current step, embeddings for three distinct inputs are required: 1) the current node (the node to be inserted), 2) unvisited nodes, and 3) the positions in the partial solution.

\paragraph{Embeddings of Positions in the Current Partial Solution } 
The unvisited node with the minimum Euclidean distance to the last node is selected as the current node to be inserted, of which the embedding is denoted as $\mathbf{h}_{u_c}$. Secondly, the embeddings of the remaining unvisited nodes can be denoted as $H_u=\{\mathbf{h}_{u_j}\}, u_j \notin \{\pi_{1:t-1}\}, u_j \neq u_c$. Two learnable linear projections $W_0\in \mathbb{R}^{d\times d}$ and $W_1\in \mathbb{R}^{d\times d}$ are applied to the current node's embedding and each unvisited node's embedding, respectively

\paragraph{Embeddings of the Current Node and Unvisited Nodes} 
To generate the embedding of a potential insertion position within the partial solution, the embeddings of adjacent visited nodes are horizontally concatenated.
For example, for a partial solution:

\begin{center}
\begin{tikzpicture}
        \node at (0,0) {\(0 \rightarrow 1 \rightarrow 2 \rightarrow 0 \rightarrow 4 \rightarrow 5 \rightarrow 6 \rightarrow 0 \rightarrow 7 \rightarrow 8 \rightarrow 0\)};
    \draw[->, semithick, black] (3.6,-0.2) -- (3.6,-0.4) -- (-3.6,-0.4) -- (-3.6,-0.2);
\end{tikzpicture}
\end{center}
where the positions are indicated by these arrows. The arrow from the tail to the head ($ 0 \leftarrow 0$) indicates the position that allows the current node to be inserted to start a new route. To simplify the notation, we assume the current incomplete solution is of length $l$, and the embeddings of the nodes in it are $ \{\mathbf{h}_i\}, i\in\{1:l\}$. Then the set of embeddings of the positions in the partial solution can be denoted as $\{\mathbf{e}_i\}, i\in\{1:l\}$, where 
\begin{equation}
\begin{aligned}
       \mathbf{e}_i &=
      \begin{cases}
		[\mathbf{h}_i, \mathbf{h}_{i+1}] &    i \in \{1: l-1\}, \\
        [\mathbf{h}_{l}, \mathbf{h}_{1}],  &    i=l,
      \end{cases}
\end{aligned}
\end{equation}
where $[\cdot,\cdot]$ is the horizontal concatenation operator. We note that both $\mathbf{h}_{1}$ and $ \mathbf{h}_{l}$ are the depot embeddings. To \textbf{incorporate the remaining capacity information}, the decoder updates each position embedding by horizontally concatenating it with the remaining capacity of the route where the position is located, increasing the embedding dimension by 1. Subsequently, a linear projection $W_2 \in \mathbb{R}^{{(2d+1)}\times d}$ is applied to the updated position embedding to project it to the embedding with the same dimension as the node embedding. The set of output position embeddings is denoted as $H_e = \{ \mathbf{e}_{\pi_i}\}, i\in\{1:t-1\}$.

Then, these embeddings together are refined by $L$ stacked attention layers. The first layer receives input $X^0=\{\mathbf{h}_{c}, H_e, H_u\}$, with each subsequent $j$-th attention layer processing the output from the $(j-1)$-th layer, culminating in the final output $X^{(L)}$. This process can be denoted as:

\begin{equation}
\label{decoder attention equation}
\begin{aligned}
        {H}^{(1)} & = \operatorname{AttnLayer}({H}^{(0)}),\\
        {H}^{(2)} & = \operatorname{AttnLayer}({H}^{(1)}),\\
        & \cdots \\
       {H}^{(L)} & = \operatorname{AttnLayer}({H}^{(L-1)}),\\
\end{aligned}
\end{equation}
Where the formulation of ``AttnLayer'' is detailed in Appendix~\ref{append:AttnLayer}. Then, a linear projection and softmax function are applied to $X^{(L)}$ to compute the insertion probabilities of the current node into all valid positions. The masking operation assigns $-\infty$ to invalid positions (those corresponding to the current node and non-visited nodes) to ensure proper probability normalization over valid insertion locations. The inserting probability $p_i$ is calculated as:
\begin{equation}
\begin{aligned}
      x_i  & = X^{(L)}_i W_f + b_f,\\
        a_{i} &= \begin{cases}
		x_i &  \forall  i \in \{1:l\} \text{ and the remaining capacity of the route where the position is located} \\
  &  \text{is not less than the current node's demand,} \\
        -\infty & \text{otherwise,} 
    \end{cases} \\
    \mathbf{p} &= \mbox{Softmax}(\mathbf{a}),
\end{aligned}
\end{equation}
where $W_f \in \mathbb{R}^{d\times 1}$ and $ b_f \in \mathbb{R}^{1}$ are learnable parameters.
where $W_f \in \mathbb{R}^{d\times 1}$ and $ b_f \in \mathbb{R}^{1}$ are learnable parameters. Each $p_i\in \mathbf{p}$ denotes the probability of inserting the current node between consecutive nodes $(\pi_i, \pi_{i+1})$ in the partial solution. The decoder executes this sampling-and-insertion procedure $n$ times to construct the complete solution $\{0,\pi_1,\cdots,0,\cdots,\pi_n\}$.

\subsection{Implementation of Insertion-based Local Reconstruction on CVRP}
In the main paper, we introduce the process of insertion-based local reconstruction on TSP. This process remains nearly the same as in CVRP, which is described as follows: 
\begin{itemize}
    \item \textbf{Solution Initialization} The initial solution is also generated by the model through a simple greedy rollout. 
    \item \textbf{Distance-based Destruction} In the CVRP case, we also destroy the solution by the node's neighbor relationship in terms of distances. Specifically, we first randomly sample a customer node. Then, we select its $\alpha$-nearest neighbor customer nodes and remove them as unvisited nodes, the left solution automatically creates an incomplete solution with multiple looped routes. Here, we refer to $\alpha$ as the destruction size. For example, if a CVRP10 solution is $(0, \pi_1, \pi_2, 0, \pi_3, \pi_4, \pi_5, 0, \pi_6, \pi_7, 0, \pi_8, \pi_9, \pi_{10}, 0)$, the node $\pi_5$ is randomly sampled, and its 3-nearest neighbors are $\{\pi_1, \pi_3, \pi_6\}$, then they are removed from the solution to serve as unvisited nodes. The left incomplete solution is $(0, \pi_2, 0, \pi_4, 0, \pi_7, 0, \pi_8, \pi_9, \pi_{10}, 0)$.
    \item \textbf{Insertion-based Reconstruction}
    The model inserts the unvisited nodes to the incomplete solution node by node. Initially, the last node is randomly selected from the incomplete solution. The subsequent procedures are the same as those for a greedy rollout. When all nodes are visited, a new complete solution is generated. We compare the previous solution with the new one, accepting the higher-quality solution for the next destruction-reconstruction cycle. This iterative process continues until the computational budget (e.g., maximum iterations or time) is reached.
    
\end{itemize}

\paragraph{Padding for Solution Misalignment} Since the solution to each CVRP instance may involve a different number of depot visits, the resulting solution lengths can vary when solving multiple CVRP instances in batches. To address this issue, we apply a simple padding technique by adding extra zeros at the end of solutions with fewer depot visits. These extra zero pairs are then masked in the model.

\section{Implementation Details of Baselines}
\label{Implementation details of Baselines}

We compare L2C-Insert with a comprehensive set of baseline methods including \textbf{(1) Classical Solvers:} Concorde~\citep{applegate2006concorde}, LKH3~\citep{LKH3}, and HGS~\cite{HGS}; \textbf{(2) Constructive NCO Methods:} POMO~\citep{kwon2020pomo}; BQ~\citep{drakulic2023bq}, LEHD~\citep{luo2023neural}, INViT~\citep{fang2024invit}, and ELG~\citep{gao2024towards}. (3) \textbf{Heatmap-based NCO Methods:} Att-GCN+MCTS~\citep{fu2021generalize}, DIFUSCO~\citep{sun2023difusco}, and T2T~\citep{li2023from}; (4) \textbf{Two-stage NCO Methods:} H-TSP~\citep{pan2023h-tsp}, SO-mixed~\citep{cheng2023select}, GLOP~\citep{ye2024glop}, and UDC~\citep{ye2024glop}; (5) \textbf{Improvement-based NCO Methods:} NeurOpt~\citep{ma2023neuopt}.

For the heatmap-based methods, we directly use their reported results. For SO-mixed, as their source codes are not released available, we choose to directly report their original results. However, we use the same test datasets for fair comparison (all these methods used the same TSP1K and TSP10K test datasets from~\citet{fu2021generalize}). For other methods, if not otherwise stated, we directly run their source codes with the default settings. 

\begin{itemize}
    \item \textbf{Conconde}~\citep{applegate2006concorde} We run the Python wrapper for the concoder, namely PyConcorde. We use the default settings across TSP instances of various sizes.
    \item \textbf{LKH3}~\citep{LKH3} We run the code provided by AM~\citep{kool2018attention}. For TSP100K, we use POPMUSIC~\citep{taillard2019popmusic} to reduce the complexity for pre-processing (by setting CANDIDATE SET TYPE = POPMUSIC and INITIAL PERIOD=1000), following the suggestion from~\citet{fu2023hierarchical}. Other settings are default.
    \item \textbf{HGS}~\citep{HGS} We run the Python wrapper for the HGS-CVRP solver, namely PyHygese.
    \item \textbf{LEHD}~\citep{luo2023neural} We limited the maximum destruction size of RRC to 1000 after observing that performance remained nearly unchanged when compared to cases without this restriction, while inference time increased significantly.
\end{itemize}

\clearpage
\newpage
\section{Hyperparameter Studies}
\label{appd: hyperparameter study}

We conduct hyperparameter studies for the values of neighborhood size $k$ and destruction size $\alpha$ by running L2C-Insert using local reconstruction with $I$=50 iterations.

\subsection{Defining the Neighbor Unvisited Nodes and Positions}
 \paragraph{Neighbor Unvisited Node} Given a current node and a unvisited node-set $B$, we categorize the unvisited nodes in $B$ that are closer to the current node as neighbor nodes based on the distance.
\paragraph{Neighbor Position} The position-node distance between the position and current node is measured by the smallest distance between the current node and the adjacent visited nodes of this position. Given a current node and a position set $E$, we categorize the positions in $E$ that are closer to the current node as neighbor positions based on the position-node distance.

\subsection{On TSP}

As detailed in Table~\ref{ablation:k-and-RRC-range-tsp}, model performance initially improves with an increase in destruction size from 100 to 300, beyond which it declines as the destruction size extends to 500. Consequently, we select an intermediate destruction size of 300 to achieve optimal performance. Furthermore, our analysis reveals that increasing the neighborhood size from 100 to 500 leads to progressively longer solution lengths. We hypothesize that the intricate interplay of numerous positions and nodes complicates the processing, thereby diminishing the model's effectiveness. Therefore, we set the neighborhood size to 100 to enable robust problem-solving while maintaining a low runtime.

\begin{table}[htbp]
\centering
\caption{Effect of the setting of neighborhood $k$ and destruction size $\alpha$ in inference stage on TSP10K.}
\resizebox{0.7\columnwidth}{!}{
  \begin{tabular}{c l | c c c}
    \toprule
    &     &    \multicolumn{3}{c}{Destruction Size $\alpha$}  \\ 
   \multicolumn{2}{c|}{}     &  100  &  300 & 500  \\ 
    \midrule
    &     &  Length (Time)  &  Length (Time) & Length (Time)  \\ 
    \midrule
    \multirow{3}{*}{Neighborhood Size $k$}
    & 100    & 76.669 (1.32m)  &   \textbf{76.573 (2.04m)}     &  77.078 (3.94m)   \\
    & 200  &  78.407 (1.68m) &   76.610 (3.11m)   & 78.320 (5.77m) \\
    & 300  &  80.349 (2.26m) &   78.983 (4.17m)   &  79.042 (7.29m)  \\
    & 500  & 81.448 (4.36m)  &   80.451 (7.48m)   &79.150 (12.18m)  \\
    \bottomrule
    \end{tabular}%
}
\label{ablation:k-and-RRC-range-tsp}
\end{table}

\subsection{On CVRP}

As indicated in Table~\ref{ablation:k-and-RRC-range-cvrp}, model performance generally improves with increasing destruction size from 100 to 300, but declines when the destruction size further extends from 300 to 500. Consequently, we select an intermediate destruction size of 300 to achieve favorable results. Furthermore, with the destruction size set at 300, the model exhibits significantly better performance when the neighborhood size is 200, leading us to choose this value for our experiments.

\begin{table}[htbp]
\centering
\caption{Effect of the setting of neighborhood $k$ and destruction size $\alpha$ in inference stage on CVRP1K.}
\resizebox{0.7\columnwidth}{!}{
  \begin{tabular}{c l | c c c}
    \toprule
     &   &    \multicolumn{3}{c}{Destruction Size $\alpha$}  \\ 
    \multicolumn{2}{c|}{}     &  100  &  300 & 500  \\ 
    \midrule
    &     &  Length (Time)  &  Length (Time) & Length (Time)  \\ 
    \midrule
    \multirow{3}{*}{Neighborhood Size $k$}
    & 100  & 43.011 (0.51m) &    43.196 (0.71m)  &  43.530 (1.31m) \\
    & 200  & 43.127 (0.95m)  &   \textbf{42.783 (1.53m)}   &  43.069 (2.78m)  \\
    & 300  & 43.030 (1.29m)  &   42.989 (2.15m)   &  43.166 (3.81m)  \\
    & 500  &  43.178 (2.53m) &   42.887 (4.37m)   &  43.204 (7.47m)  \\
    \bottomrule
    \end{tabular}%
}
\label{ablation:k-and-RRC-range-cvrp}
\end{table}

\clearpage
\newpage
\section{Additional Ablation Studies and Analysis}
\label{Additional Ablation Studies}
Each step in the insertion-based construction process is divided into two substeps: (1) selecting a node from the unvisited nodes and (2) selecting an edge in the incomplete solution for insertion. In this paper, the second substep is conducted by the model, and the first substep is operated through predetermined rules. Specifically, for the first substep, we define the insertion node selected in the previous step as the last node, and the insertion node selected in the current step as the current node. The current node is the unvisited node closest to the last node. In this section, we study the impact of the distance criterion and the current node selection strategy on model performance. Beyond those, we also study the impact of adding the Positional Encoding~\citep{vaswani2017attention}
to the embeddings of the positions in the partial solution, the effect of unvisited nodes as decoder input, and the performance of L2C-Insert using different backbone models.

\subsection{Effect of Distance Criterion for CVRP}

We test two classic distance criteria, Euclidean Distance and Polar Angle. For two customer nodes  $\mathbf{p} = (p_1, p_2)$ and $\mathbf{q} = (q_1, q_2)$, Euclidean Distance is defined as:$
d_E(\mathbf{p}, \mathbf{q}) = \sqrt{ (p_1 - q_1)^2 + (p_2 - q_2)^2}$. For Polar Angle, we must introduce the depot node $\mathbf{d}_c = (d_1, d_2)$ to serve as the Polar center. Then it is defined as: 
$
\theta_P(\mathbf{p}, \mathbf{q}) = \left|\theta_{\mathbf{p}} - \theta_{\mathbf{q}}\right|
$,
where $\theta_{\mathbf{p}} = \arctan\left(\frac{p_2 - d_2}{p_1 - d_1}\right)$ and $\theta_{\mathbf{q}} = \arctan\left(\frac{q_2 - d_2}{q_1 - d_1}\right)$ are the angles formed by the lines connecting the depot $\mathbf{d}_c$ to the customer nodes $\mathbf{p}$ and $\mathbf{q}$ with respect to the horizontal axis.

We train and test the same model with these two distance criteria, maintaining the node selection strategy that the current node is the closest unvisited node to the last one. Both models are trained with the same budget mentioned in the main paper. We then test these models on CVRP100 and CVRP1000 using a greedy rollout. The results, as shown in Table~\ref{Effect of distance criteria for CVRP}, indicate that the model trained and tested with Polar Angle performs much better. This improvement can be attributed to the nature of CVRP, where multiple routes are usually ordered around a central depot. Using polar angles is tightly coupled to the solution structure of CVRP. As a result, the model is easier to learn and ultimately achieves better performance.
Therefore, we use the Polar Angle to measure the neighbor relationship on CVRP.

\begin{table}[htbp]
\centering
\caption{Effect of distance criterion for CVRP.}
\resizebox{0.4\columnwidth}{!}{
\begin{tabular}{c  | c c }
\toprule
 & CVRP100  & CVRP1000       \\
 &  Gap  &  Gap \\ 
\midrule
 Euclidean Distance& 5.151\%  & 10.155\%     \\ 
  Polar Angle & \textbf{3.892\%}  &  \textbf{8.009\%}     \\   
\bottomrule
\end{tabular}%
}
\label{Effect of distance criteria for CVRP}
\end{table}

\subsection{Effect of Node Select Strategy}
\label{Effect of Node Select Strategy}
We use two different current node selection strategies described in~\citet{nilsson2003heuristics} for comparison: 

\begin{itemize}
    \item Random: select an unvisited node randomly.
    \item Nearest Neighbor: select the unvisited node that is nearest to the last node.
\end{itemize}

Specifically, we train the same models with two different current node selection strategies. The CVRP models use the same training budget mentioned in the main paper, while TSP models are trained with a budget of 10 epochs due to the runtime. We test these models on TSP100 and CVRP100 using greedy rollout. The results in Table~\ref{Effect of selection strategy} show that the nearest neighbor node selection strategy has the best overall result. Therefore, in this paper, we choose to use it for the selection of the current node.

\begin{table}[h]
\centering
\caption{Effect of selection strategy.}
\resizebox{0.4\columnwidth}{!}{
\begin{tabular}{c  | c   | cc}
\toprule
 & TSP100     & CVRP100       \\
 &  Gap     &  Gap  \\ 
\midrule

Random           & 1.1557\%   &   11.370\% \\
Nearest Neighbor & \textbf{ 0.9535\%}  &  \textbf{3.892\%}  \\ 
\bottomrule
\end{tabular}
}
\label{Effect of selection strategy}
\end{table}

\subsection{Effect of Positional Encoding}
\label{Effect of Positional Encoding}
We compare the performance of models with and without incorporating the Positional Encoding (PE) from ~\citep{vaswani2017attention} to the embeddings of positions in the partial solution. The models are trained on TSP100 with a budget of 10 epochs, and tested on TSP100 using greedy rollout. The results in Table~\ref{Effect of PE} show that incorporating PE will damage both in-domain and out-of-domain performance. Therefore, this paper does not include the positional encoding in our model.

\begin{table}[h]
\centering
\caption{Effect of selection strategy.}
\resizebox{0.33\columnwidth}{!}{
\begin{tabular}{c  | c   | cc}
\toprule
 & TSP100     & CVRP100       \\
 &  Gap     &  Gap  \\ 
\midrule

W/  PE          & 1.0475\%   &   5.802\% \\
W/o PE & \textbf{ 0.9535\%}  &  \textbf{3.892\%}  \\ 
\bottomrule
\end{tabular}%
}
\label{Effect of PE}
\end{table}

\subsection{Effect of Decoder Input of Unvisited Nodes}
\label{Effect of Model Input of Unvisited Nodes}
We compare the performance of models trained on TSP100 with and without inputting unvisited nodes to the decoder. The results in table~\ref{abla:train without unvisited node} show that while inputting unvisited nodes to the model results in additional computational cost, the performance improvement is significant, proving its indispensability.

\begin{table}[ht]
\centering
\caption{Training with unvisited node vs. without.}
\label{abla:train without unvisited node} 
    \resizebox{0.4\textwidth}{!}{
\begin{tabular}{c  | c c | c c }
\toprule[0.5mm]
&   \multicolumn{2}{c|}{TSP100}   &  \multicolumn{2}{c}{TSP1K}       \\
Method &  Gap & Time  &  Gap & Time  \\ 
\midrule
W/o   & 5.161\% &  0.44m &  23.979\%  &  0.15m   \\
W/ & \textbf{0.460\%}  & 0.86m  & \textbf{4.757\%} &  0.36m    \\
\bottomrule[0.5mm]
\end{tabular}%
    }
\end{table}

\paragraph{Why the L2C-Insert Model Necessitates the Unvisited Nodes as the Decoder Input}

\begin{figure}[ht]
  \centering
  \includegraphics[width=0.35\linewidth]{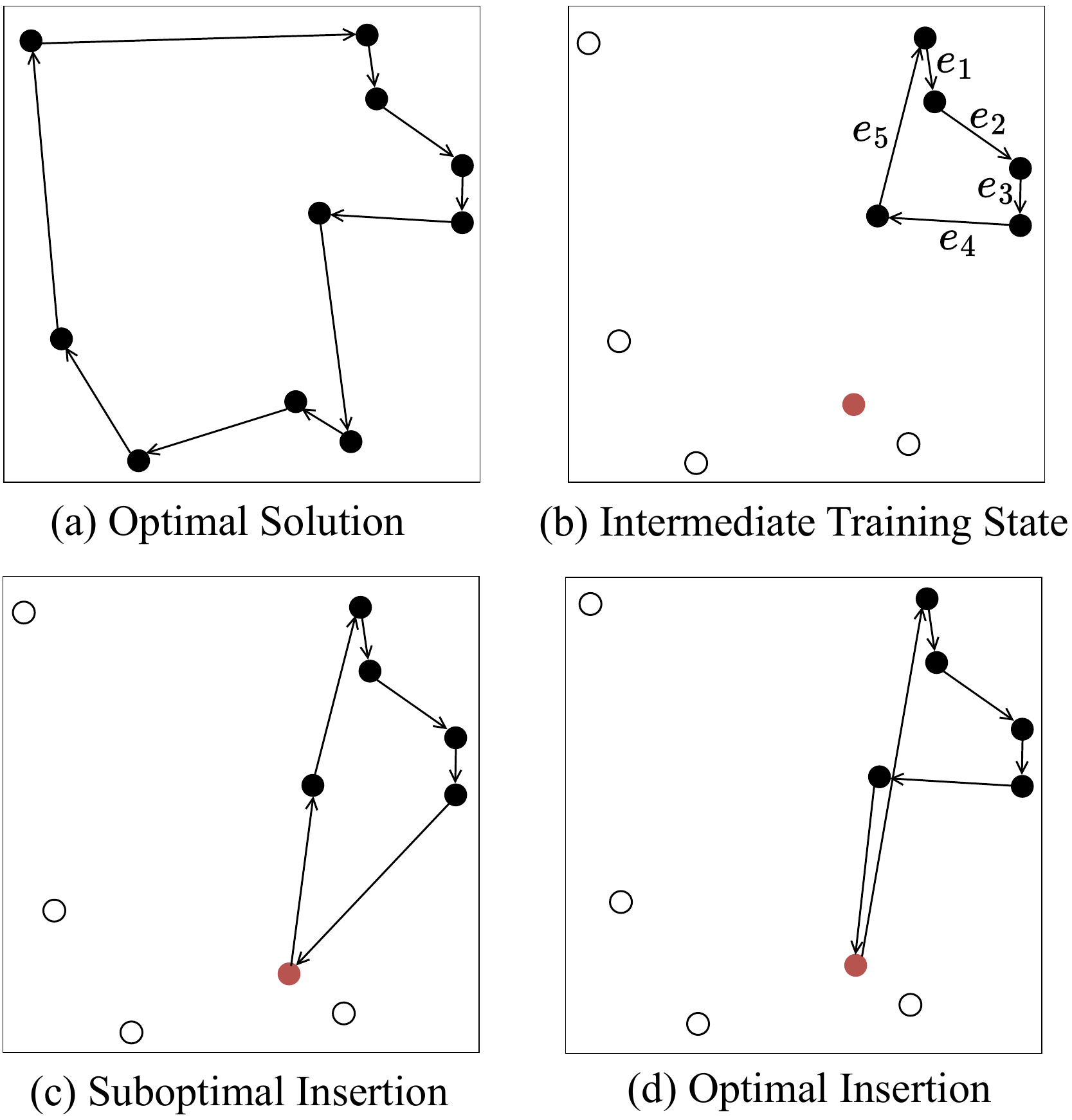} 
  \caption{figure}{Example of a suboptimal insertion (not considering unvisited nodes) versus an optimal insertion (considering unvisited nodes). The red node is the current node to insert.}
  \label{discussion_model_input_unvisited_node}
\end{figure}

To illustrate why this input is crucial, consider the example depicted in Figure~\ref{discussion_model_input_unvisited_node}. Figure~\ref{discussion_model_input_unvisited_node} (a) displays the optimal solution for a TSP instance, while Figure~\ref{discussion_model_input_unvisited_node} (b) shows an intermediate state during the solution construction process. Without considering unvisited nodes, the insertion action shown in Figure~\ref{discussion_model_input_unvisited_node} (c), despite yielding a better partial solution, constitutes a suboptimal choice when evaluated against the global optimum considering all nodes. In contrast, when unvisited nodes are factored in, the insertion action in Figure~\ref{discussion_model_input_unvisited_node} (d), even though it results in a partial solution with path intersections, is identifiable as a more globally optimal insertion by referencing the overall optimal solution. Therefore, insertions made without information about unvisited nodes tend to be shortsighted. Incorporating unvisited nodes, however, provides the model with a broader context, enabling more informed decisions for selecting insertion positions. Consequently, the L2C-Insert model needs to leverage information from unvisited nodes to mitigate such shortsightedness, thereby guiding the construction process towards higher-quality solutions.

\subsection{Performance of L2C-Insert using Different Backbone Models}
We conduct experiments to investigate the performance of the proposed L2C-Insert framework under different backbone NCO models. Specifically, three recently proposed models are employed: POMO~\citep{kwon2020pomo}, LEHD~\citep{luo2023neural}, and INVIT~\citep{fang2024invit}, which feature 1, 3, and 6 decoder layers, respectively, each within a 7-layer overall model architecture. These models represent light, balanced, and heavy decoder architectures, respectively. Building upon these models, L2C-Insert variants are developed using the proposed L2C-Insert framework. 
These models are evaluated using the TSPLIB and CVRPLIB instances detailed in Section~\ref{Section: Experiment}. Both the L2C-Append and L2C-Insert methods utilize a one-shot greedy search for inference. 

The experimental results are presented in Table~\ref{Append vs. Insert, Learning}. 
These results indicate that for the LEHD model, the L2C-Insert method outperforms the L2C-Append method on 4 out of 6 instance groups. Conversely, the L2C-Insert method, when based on the POMO model, generally underperforms the L2C-Append method, proving inferior in 3 out of 6 instance groups, respectively. Similarly, the L2C-Insert method, built upon the INViT model, also generally underperforms L2C-Append, showing inferiority in 6 out of 6 instance groups.
These findings suggest that the L2C-Insert method may be more effective when utilized with a heavy decoder model. This advantage could stem from the complex position relationships within an incomplete solution and the current node being inserted during solution construction. Capturing these intricate relationships effectively necessitates a heavy decoder with sufficient model capacity to inform appropriate decisions regarding the correct insertion position.

\begin{table}[ht]
  \centering
  \caption{Performance of L2C-Insert and L2C-Append using different backbone models. These methods use the greedy rollout (Greedy) for inference.}
    \resizebox{\textwidth}{!}{%
    \begin{tabular}{ll|ccc|ccc|c}
    \toprule
    \multicolumn{2}{l|}{Method } & \multicolumn{3}{c|}{\textbf{TSPLIB }} & \multicolumn{3}{c|}{\textbf{CVRPLIB }} & \multirow{2}[4]{*}{Perform Better \# } \\
\cmidrule{3-8}    \multicolumn{2}{l|}{(Using Greedy)} & N<=200 & 200<N<=1K & All & N<=200 & 200<N<=1K & All &  \\
    \midrule
    POMO  & Append & \textbf{5.325\%} & \textbf{35.337\%} & \textbf{17.575\%} & 15.596\% & 41.759\% & 36.003\% & 3/6 \\
          & Insert & 19.111\% & 38.344\% & 26.961\% & \textbf{14.650\%} & \textbf{19.305\%} & \textbf{18.281\%} & 3/6 \\
    \midrule
    \midrule
    INViT & Append & \textbf{2.788\%} & \textbf{6.315\%} & \textbf{4.228\%} & \textbf{8.046\%} & \textbf{11.313\%} & \textbf{10.594\%} & 6/6 \\
          & Insert & 3.961\% & 8.107\% & 5.653\% & 14.927\% & 22.551\% & 20.874\% & 0/6 \\
    \midrule
    \midrule
    LEHD  & Append & 1.922\% & \textbf{3.646\%} & \textbf{2.626\%} & 11.346\% & 12.851\% & 12.520\% & 1/6 \\
          & Insert & \textbf{1.526\%} & 4.382\% & 2.691\% & \textbf{10.903\%} & \textbf{9.548\%} & \textbf{9.846\%} & 4/6 \\
    \bottomrule

    \end{tabular}
    }
  \label{Append vs. Insert, Learning}
\end{table}%

\section{Discussion on the Advantage of the Insertion-based Construction Paradigm in Search}

The iterative local reconstruction is a significant factor in achieving the final reported performance. We believe the key innovation of our work, the insertion-based policy, is what unlocks the true potential of this search procedure. The policy's inherent flexibility is a key enabler for the search's effectiveness.

\textbf{1. The advantage of insertion-based search}

As we discuss in Section~\ref{subsection:Insertion-based Local Reconstruction}, conventional appending-based models face a fundamental limitation when used with local search. When nodes are removed during the destruction phase, they can only be re-appended to the end of the sequence during reconstruction. This rigidity makes it difficult for the model to escape local optima. For instance, reconnecting two nodes that are geometrically close but far apart in the sequence—a common requirement for improving a tour—is highly improbable.

Our L2C-Insert framework directly overcomes this bottleneck. By allowing a node to be re-inserted at any valid position, our model facilitates a much more powerful and effective exploration of the solution space during the reconstruction phase. This flexibility is the core reason why the destroy-and-reconstruct loop is effective in our framework.

\textbf{2. Direct Ablation Study: Isolating the Policy's Contribution}

To provide a direct and fair comparison that disentangles the policy's contribution, we conduct the ablation study presented in Table~\ref{abla: L2C-Append vs. L2C-Insert} in the main paper. In this experiment, we compared our L2C-Insert model against an L2C-Append model (which uses a similar architecture but an appending policy) within the exact same local reconstruction framework.

The results show that our L2C-Insert with only 200 iterations significantly outperforms the L2C-Append model that uses 1000 iterations. This finding directly demonstrates that the superior performance stems from the intrinsic quality and flexibility of the insertion policy itself, and not just from the application of a search method.

\section{Clarification on Fair Comparison and Calculating Costs}

\textbf{1. Clarifying Iteration Settings of Baselines}

To clarify, we summarize the iteration settings of iterative methods in the table below. From this table, we can observe that he number of iterations varies significantly. For instance, NeurOpt utilizes 10,000 lightweight k-opt iterations, while two-stage methods like UDC and SO-mixed perform 50-100 divide-and-conquer iterations.

\begin{table}[htbp]
  \centering
  \caption{Iteration settings for baselilne methods. "Reviser-$n$" refers to a solver designed to repair a local solution of length $n$.}
     \resizebox{0.5\textwidth}{!}{
    \begin{tabular}{c|c|c}
    \toprule
    Method & Problem & Number of iteration  \\
        \midrule
  NeurOpt & TSP \& CVRP  & 10000 (k-opt)\\
    \midrule
   SO-mixed & TSP1K & 50 (Reviser-100/50)\\
  & TSP10K & 100 (Reviser-100/50)\\
    \midrule
    UDC & TSP \& CVRP   & 50 for Reviser-100 (with 50 initial solutions) \\
    \midrule
    GLOP  & TSP  & 100/50/10 for Reviser-100/50/20\\
      & CVRP    & 5 for LKH3 as Reviser-20\\
    \bottomrule
    \end{tabular}%
    }
  \label{tab:addlabel}
\end{table}

\textbf{2. A Note on Comparing Computational Budgets}

We would like to respectfully clarify that directly comparing 'iteration counts' may be challenging, as the definition and computational cost of an "iteration" differ substantially across methods:
\begin{itemize}
    \item Improvement-based methods like NeurOpt perform very fast, local search operations (e.g., k-opt) in each iteration.
    \item Two-stage methods like GLOP and UDC involve multiple sub-problem solvers per iteration or perform the destroy-and-repair on multiple initial solutions, and also may call a powerful heuristic solver like LKH3 internally.
    \item Our L2C-Insert performs local reconstruction only on a single solution using one neural model.
\end{itemize}
For this reason, the total inference time may offer a more holistic and equitable basis for comparing the computational budget. Our analysis is therefore centered on the solution quality (optimality gap) achieved within a given runtime. Our experimental results, when viewed in terms of total runtime, seem to support this perspective. For example, in Table 1 (TSP1K), our L2C-Insert (I=1000) achieves a 0.481\% gap in 46.4 minutes. In contrast, a strong baseline like LEHD+RRC1000 requires 5.2 hours to achieve a 0.731\% gap. This suggests that our method appears to offer a favorable trade-off between time and solution quality.

\section{Total Inference Time and Average Inference Time Per Instance}

The apparent decrease in total inference time for larger instances is due to the different number of instances in the test datasets for each problem size.

Following the standard evaluation protocols established in prior NCO research~\citep{luo2023neural,fu2021generalize}, the test sets for larger problems contain significantly fewer instances. For example, our TSP100 test set contains 10,000 instances, while the TSP10K and TSP100K sets contain only 16 instances each. The total inference times reported in Tables 1 and 2 of our main paper reflect the time required to solve the entire dataset.

To provide a clearer picture, the table below breaks down the total inference time into the average time per instance.

\begin{table}[htbp]
  \centering
  \caption{Add caption}
     \resizebox{0.99\textwidth}{!}{
    \begin{tabular}{l|l|l|l|l}
    \toprule
          & TSP100 & TSP1K  & TSP10K & TSP100K \\
    Dataset Size & 10000 instances & 128 instances & 16 instances & 16 instances \\
    \midrule
    L2C-Insert Total Inference Time & 8.0h  & 46.4m & 16.64m & 26.07m \\
    L2C-Insert Inference Time / instance & 2.9s  & 21.8s & 62.4s & 97.8s \\
    \bottomrule
    \end{tabular}%
    }
  \label{tab:R1-Q1}%
\end{table}%

As the table clearly shows, the average inference time per instance for L2C-Insert scales with the problem size as expected, increasing from 2.9 seconds for TSP100 to 97.8 seconds for TSP100K.

\newpage

\section{Clarification on Method Positioning and Technical Contribution}

Regarding the positioning of our method (construction heuristic vs. improvement heuristic) and its technical contributions, we make the following discussion to clarify and strengthen the core value of our paper.

The "insertion-based construction" is a well-established concept in the traditional field of Operations Research. The core contribution of our work is not the invention of the 'insertion' operation itself, but rather being the first learning-driven, insertion-based framework specifically designed for solution construction in the field of NCO for solving the vehicle routing problem, which is currently dominated by the 'appending-based' paradigm.

The make the technical contribution of this paper clear, we elaborate from the following three perspectives:

\textbf{1. The Essence of Our Core Method: A Pure Insertion-based Construction Method}

Our core model, L2C-Insert, is fundamentally a construction-based model in its design. Its task is to learn a policy $p$(position∣node, partial tour) to predict the best insertion position, given an incomplete solution (partial solution) and a node to be inserted. This process starts from an initial solution (e.g., one node) and builds a complete solution by sequentially adding unvisited nodes, which fully aligns with the definition of a construction-based neural model.
\begin{itemize}
    \item \textbf{Model Architecture and Training:} As described in Sections 3.1 and 3.2 of the paper, our model architecture (especially the decoder) and the supervised learning scheme are designed to learn this single-step constructive action. The model learns to master the optimal "insertion" decision at each step by reconstructing a given optimal solution. This is fundamentally different from an improvement heuristic (like 2-opt), which requires a complete solution as input for iterative refinement.
    \item \textbf{Comparison with the Appending-based Paradigm:} The starting point of our work is precisely to overcome the limitations of the popular "appending-based paradigm" in existing NCO methods (as mentioned in the Introduction and Figure 1). "Appending" is a construction method, and so is "insertion." However, the latter offers greater flexibility and a larger solution space, which is the core of our exploration.
\end{itemize}

\textbf{2. The Application of the Method: Distinguishing the Core Construction Model from the Inference Improvement Strategy} 

We would like to clarify that our adoption of the Insertion-based Local Reconstruction strategy during the inference stage (Section 3.3) is an advanced search strategy that uses our learned "construction model" as a core operator, rather than being the model itself.

This can be understood as a two-tiered structure:

\begin{itemize}
    \item \textbf{Lower Level (Core Contribution):} A learned, efficient construction heuristic model (L2C-Insert) that knows how to intelligently "insert" nodes.
    \item \textbf{Upper Level (Application Strategy):} An improvement framework (akin to Large Neighborhood Search, LNS) that utilizes the lower-level construction model to iteratively "destroy and repair" a solution during inference to enhance its quality.
\end{itemize}

This practice of "embedding a learned construction model into an improvement framework" is one of the most recognized, state-of-the-art, and effective practices in the current NCO field (as referenced in the paper, e.g., LEHD~\citep{luo2023neural}, GLOP~\citep{ye2024glop}). Our storyline does not confuse the two. Instead, it is:

\begin{itemize}
    \item We propose a new, more flexible construction model (L2C-Insert).
    \item To validate its powerful performance, we follow the best practices in the field and use it as a potent "local reconstruction" tool during the inference stage.
\end{itemize}

\textbf{3. Clear Technical Contributions}

Based on the two points above, our technical contributions can be summarized more clearly as follows:

\begin{itemize}
    \item \textbf{Paradigm Innovation:} In the NCO field, we are the first to explore and successfully implement a learning-driven "insertion-based" construction framework, breaking the reliance of previous research on "appending-based" construction. This is an important supplement and a rethinking of the existing NCO construction methodology.
    \item \textbf{Novel Model Architecture and Training Mechanism:} We designed a specific decoder structure (Section 3.1) to handle the complex "insertion" decision. It needs to simultaneously understand the intricate relationships between the node to be inserted, all unvisited nodes, and all possible insertion positions in the current partial solution. This is much more complex than an "appending" model, which only needs to focus on the last node. The corresponding supervised learning training scheme (Section 3.2) is also newly designed to fit the "insertion" task.
    \item \textbf{Synergistic Inference Strategy:} We propose a distance-based destruction strategy (Section 3.3) that is highly synergistic with the "insertion" paradigm. As described in the paper, this strategy can more effectively identify and break "bad edges" that are spatially close but sequentially distant, a flexibility that traditional "sequence-based destruction strategies" lack. The experiment (Table 5) also proves its effectiveness. This demonstrates the powerful synergy between our proposed construction paradigm and inference strategy.
\end{itemize}

In summary, our work is not a simple application of a known "insertion heuristic" to NCO. Our core is a novel construction model, and its application during the inference stage is an advanced and reasonable improvement strategy. We believe that this clear hierarchical division and the technical innovations at each level together constitute the solid technical contribution of this paper.

\newpage

\section{Analysis of Computational Complexity for Insertion-based and Appending-based heavy decoder-based NCO model}

A formal analysis of computational complexity is essential for a thorough comparison. Here is a detailed comparison:

\textbf{Computational Complexity of Appending-based NCO (L2C-Append)}
 
The primary source of computational complexity of the Transformer-based model comes from the attention mechanism employed. 

In a heavy decoder-based appending-based NCO model (e.g., LEHD, where heavy decoder means the decoder has six attention layers), the task at each construction step $t$ is to select one node from the $n - t$ unvisited nodes. The decoder computes a context query and attends to the embeddings of all $n$ initial nodes to determine the next node to append. This attention mechanism has a complexity of $O((n-t)^2 \cdot d)$, where $n$ is the total number of nodes and $d$ is the embedding dimension. This process is repeated for $n$ steps. Thus, the total complexity for constructing a single solution is $O(\frac{n^3}{3} \cdot d)$. In Big O notation, this is simplified to $O(n^3⋅d)$.

\textbf{Computational Complexity of Insertion-based NCO (L2C-Insert)}

Our proposed L2C-Insert framework also adopts a heavy decoder-based model. The task at each step $t$ is more complex. To make an informed decision, our decoder processes a much richer set of inputs simultaneously: the embedding of the current node to be inserted, the embeddings of all $n − t$ remaining unvisited nodes, and the embeddings of all $t − 1$ possible insertion positions. The total number of input embeddings processed by the self-attention layers at every step is therefore $(1)+(n - t)+(t - 1)=n$. The complexity of a single construction step is $O(n^2⋅d)$. This step is repeated $n$ times to build the full solution. Consequently, the total theoretical complexity is $O(n^3⋅d)$.

\textbf{Discussion and Practical Implications}

Theoretically, both heavy-decoder paradigms exhibit a cubic complexity of $O(n^3⋅d)$. However, the insertion-based paradigm is computationally more intensive as it consistently processes $n$ nodes per step, whereas the appending-based approach processes a decreasing number of nodes $(n-t)$. This higher computational cost is a direct trade-off for the enhanced flexibility and expressive power of the insertion mechanism.

To ensure practical applicability to large-scale problems, we can employ a k-nearest neighbor (k-NN) strategy during inference. Instead of processing all $n$ contextual nodes, the decoder's attention mechanism is constrained to only the k-nearest unvisited nodes and positions. This reduces the effective sequence length from $n$ to a much smaller, constant $k$ (e.g., $k=100$ in our paper), making the per-step complexity approximately $O(k^2⋅d)$.

\newpage

\section{Clarification on the Definition of Target Position}

We provide the following clarification to clarify the definition of the target position.

\textbf{1. Why two visited nodes in the current partial solution determine the target position}

During the process of the insertion construction, at each step, an unvisited node is inserted into a position of the partial solution constructed in the current step. Since 1) the nodes connected in the current partial solution are all visited nodes; 2) an insertion position is essentially an edge and is uniquely determined by the two nodes it connects, therefore, the only insertion position is determined by two visited points.

\textbf{2. How the target position is determined }

To make the process of identifying the target position clearer, we break it down into two steps. Given a labeled solution, a current partial solution, and a selected unvisited node to insert:

\textbf{Step 1: Identify the circular sequence in the labeled solution}

We first consider the complete labeled solution $\boldsymbol{\pi}^*$ and remove all unvisited nodes except for the currently selected one. This leaves us with a circular sequence containing only the selected unvisited node and the nodes already in the partial solution, preserving their original ordering from $\boldsymbol{\pi}^*$.

For example, suppose the labeled solution is $\boldsymbol{\pi}^*=(\pi_1, \pi_2, \pi_3, \pi_4, \pi_5, \pi_6, \pi_7)$, the partial solution is $(\pi_1, \pi_3, \pi_6, \pi_7)$, and the selected unvisited node is $\pi_2$. Then the other unvisited nodes are $\{ \pi_4, \pi_5\}$. After removing $\{ \pi_4, \pi_5\}$ from $\boldsymbol{\pi}^*$, the resulting circular sequence is $(\pi_1, \pi_2, \pi_3, \pi_6, \pi_7)$.

\textbf{Step 2: Identify the target position in the current partial solution}

Within this newly formed circular sequence, we can find the two nodes that are now adjacent to the selected unvisited node. These two nodes, which are part of the partial solution, define the target position for the insertion.

Continue the above example. In the circular sequence $(\pi_1, \pi_2, \pi_3, \pi_6, \pi_7)$, the nodes adjacent to $\pi_2$ are $\{\pi_1,\pi_3\}$. Since $\pi_1$ and $\pi_3$ are connected in the current partial solution, these two visited nodes determine the target position.

\newpage

\section{Rationale and Principles of the Logarithmic Loss Function in Equation~\ref{eq: loss function}}

The use of the negative log-likelihood loss function in Equation~\ref{eq: loss function} is a standard and principled choice for our supervised learning framework, primarily because it is highly effective for gradient-based optimization.

Our model treats the selection of an insertion position at each construction step as a classification problem. The model's decoder, via a softmax function, outputs a probability distribution $p_{\theta}$ over all valid insertion positions. Our objective is to train the model to assign the highest possible probability to the correct target position $\vec{e}^{\,*}$.

The logarithmic nature of this loss function $\mathcal{L}(\boldsymbol{\theta}) =  - \log p_{\boldsymbol{\theta}}\left(\vec{e}^{\,*}  \mid  \vec{e}_{\pi_{1: l}},\boldsymbol{\pi}^*\right)$ provides two key advantages for training:

\textbf{Strong Error Correction:} When the model assigns a very low probability to the correct target position (i.e., $p_{\theta}(\vec{e}^{\,*})\rightarrow$ 0 ), the loss $-log \ p_{\theta}(\vec{e}^{\,*})$ approaches infinity. This results in a very large gradient, providing a strong signal to the optimizer to correct the model's parameters significantly.

\textbf{Stable Convergence:} Conversely, as the model's prediction for the correct class approaches 1, the loss approaches 0. This leads to smaller gradients, promoting stable convergence once the model learns to make correct predictions. This behavior effectively focuses the training effort on the most difficult examples.

\clearpage
\newpage

\section{Visualization}
\begin{figure}[htbp]
	\centering
	\begin{minipage}{0.19\linewidth}
		\centering
		\includegraphics[width=0.9\linewidth]{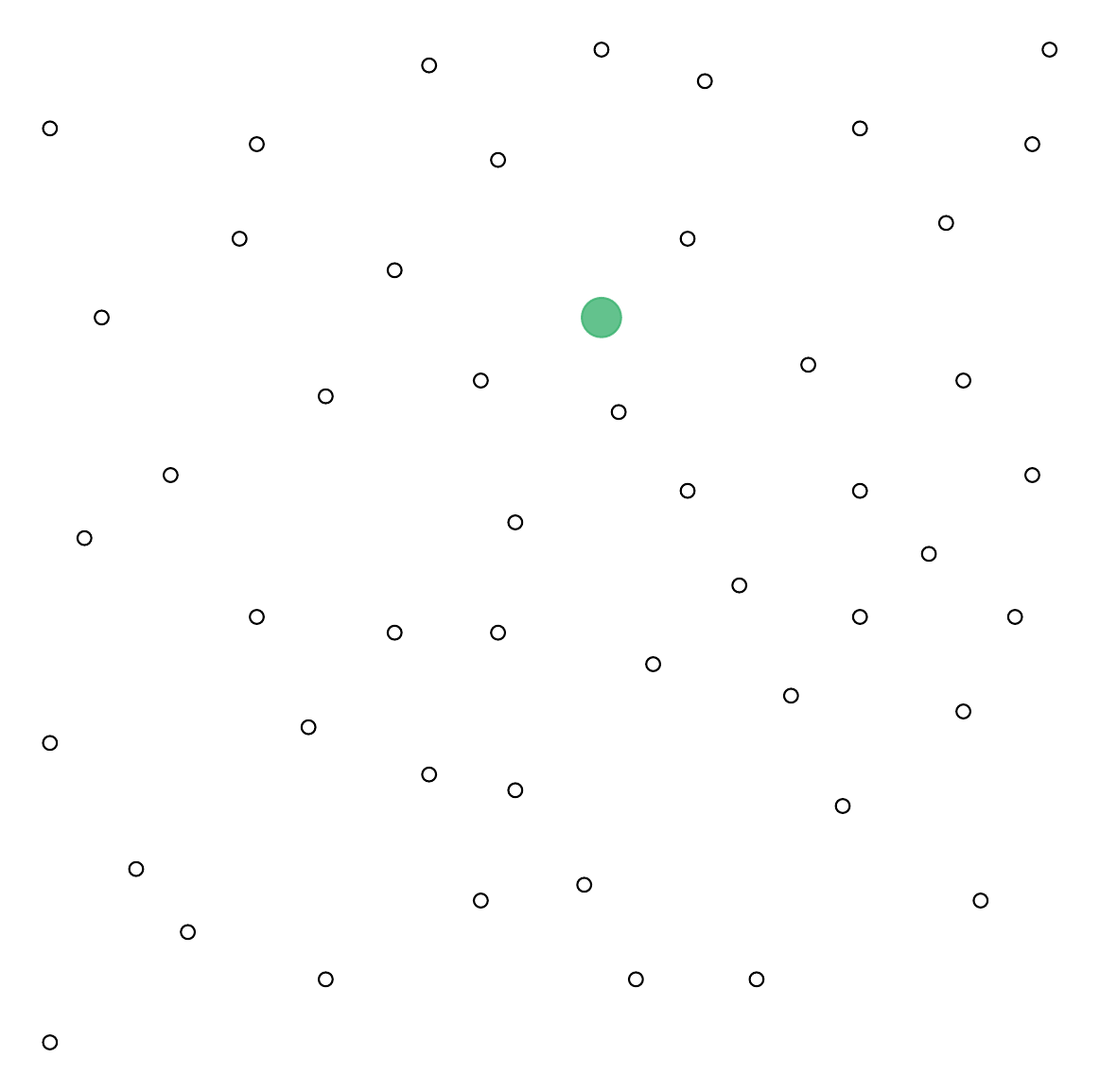}
		\text{Step 1}
	\end{minipage}
	\begin{minipage}{0.19\linewidth}
		\centering
		\includegraphics[width=0.9\linewidth]{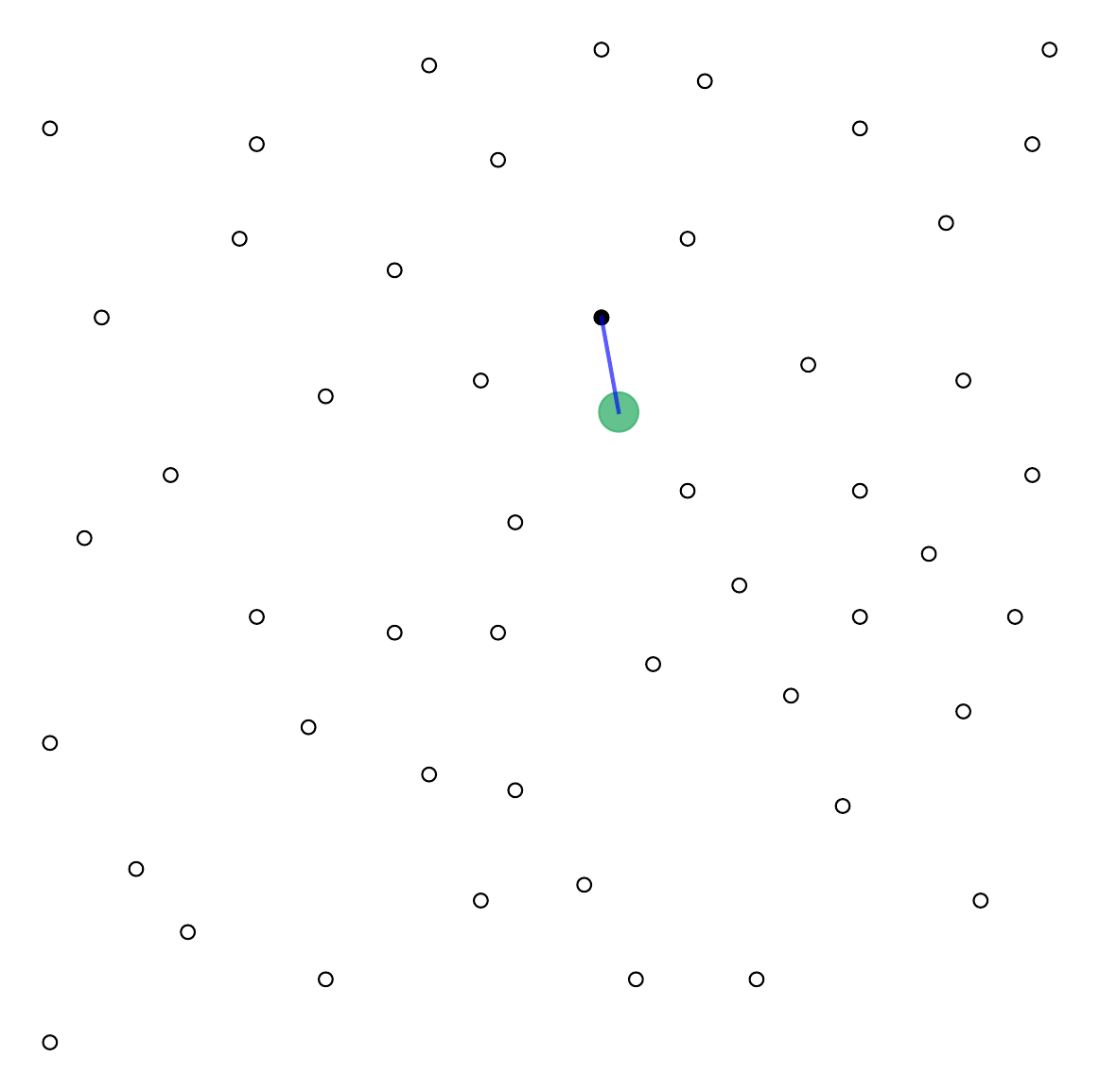}
		\text{Step 2}
	\end{minipage}
 	\begin{minipage}{0.19\linewidth}
		\centering
		\includegraphics[width=0.9\linewidth]{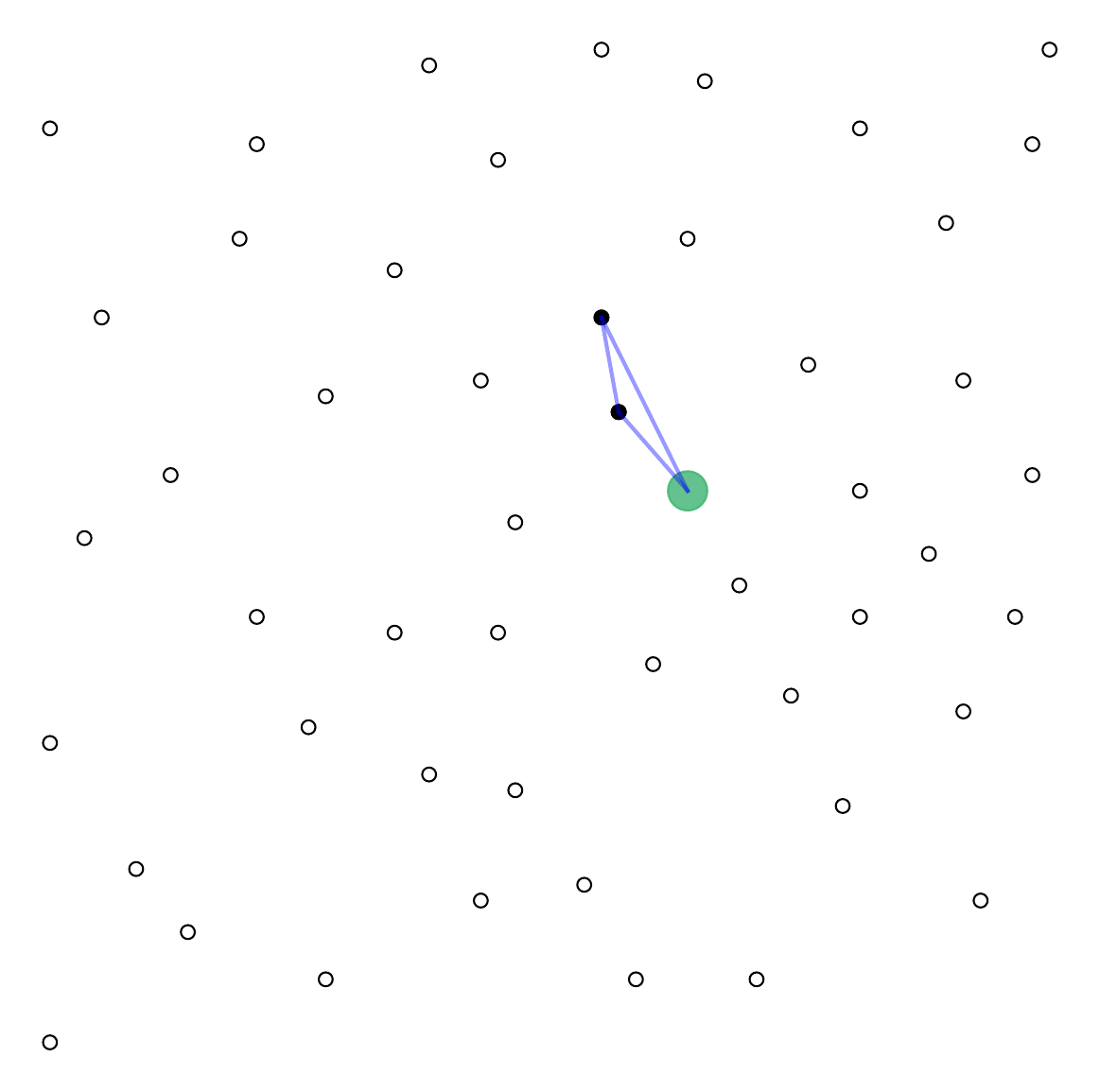}
		\text{Step 3}
		\label{f3}
	\end{minipage}
	\begin{minipage}{0.19\linewidth}
		\centering
  	\text{......}
	\end{minipage}
 	\begin{minipage}{0.19\linewidth}
		\centering
		\includegraphics[width=0.9\linewidth]{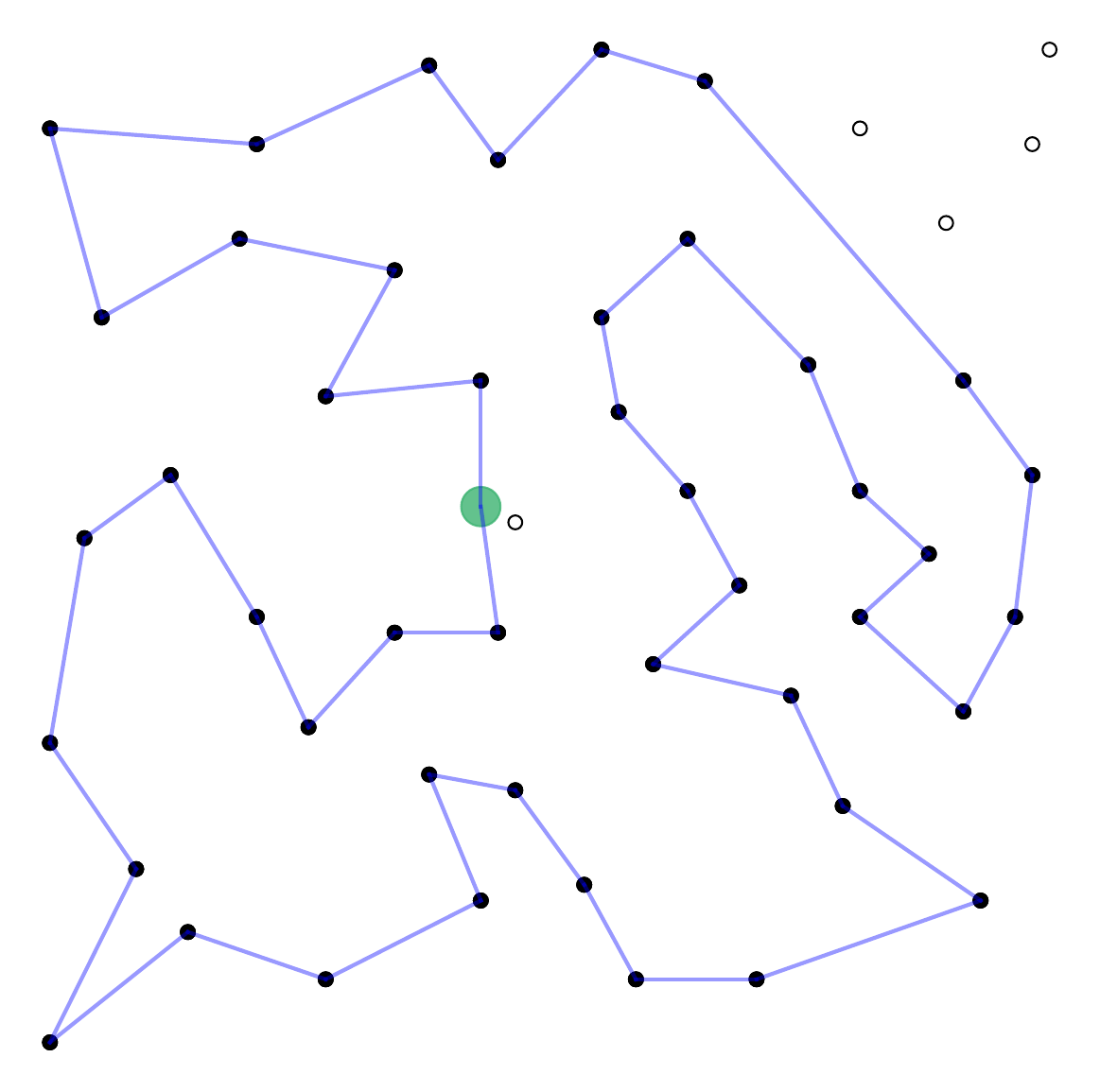}
		\text{Step 46}
	\end{minipage}
	\qquad
	
	\begin{minipage}{0.19\linewidth}
		\centering
		\includegraphics[width=0.9\linewidth]{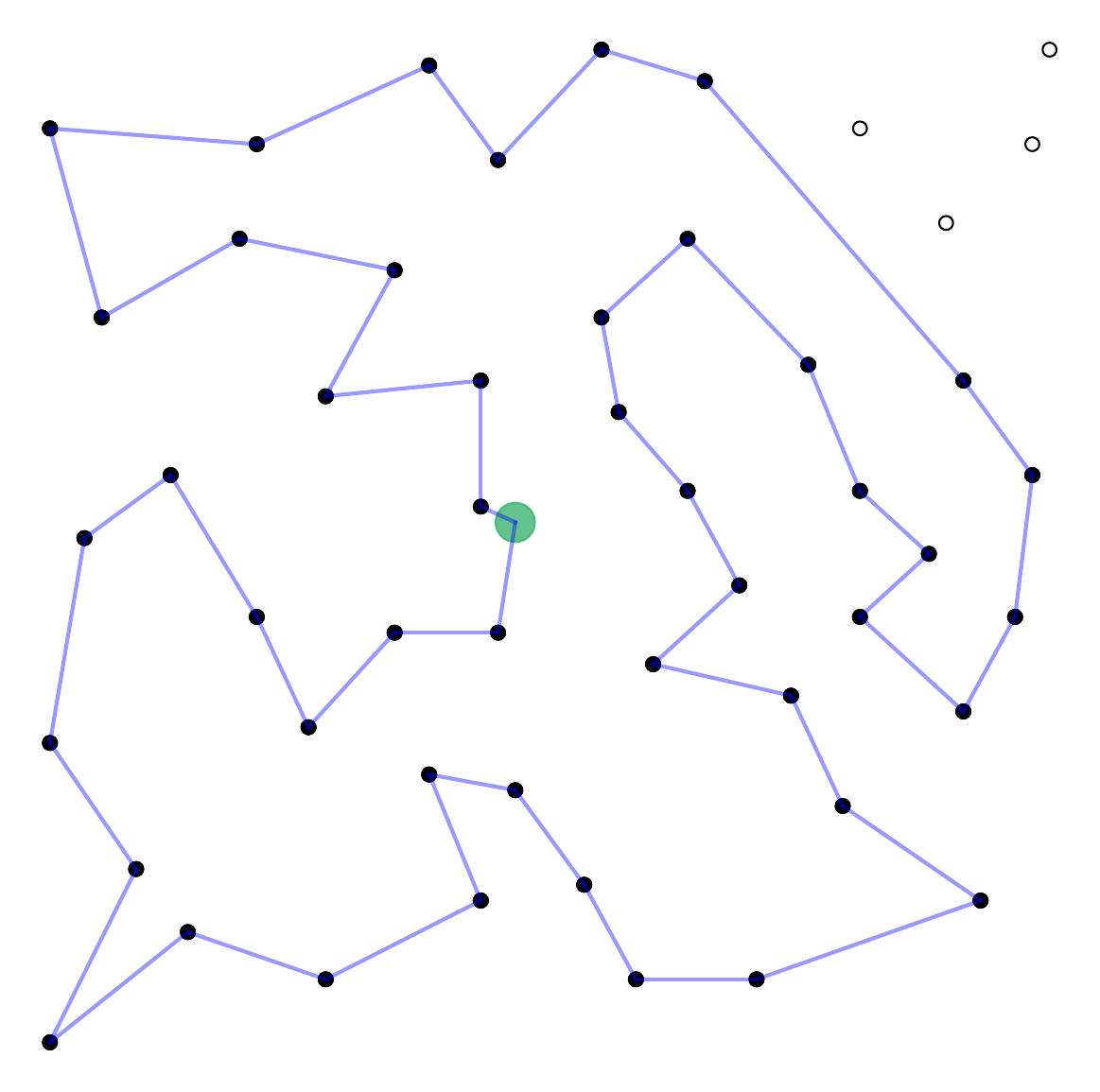}
		\text{Step 47}
	\end{minipage}
	\begin{minipage}{0.19\linewidth}
		\centering
		\includegraphics[width=0.9\linewidth]{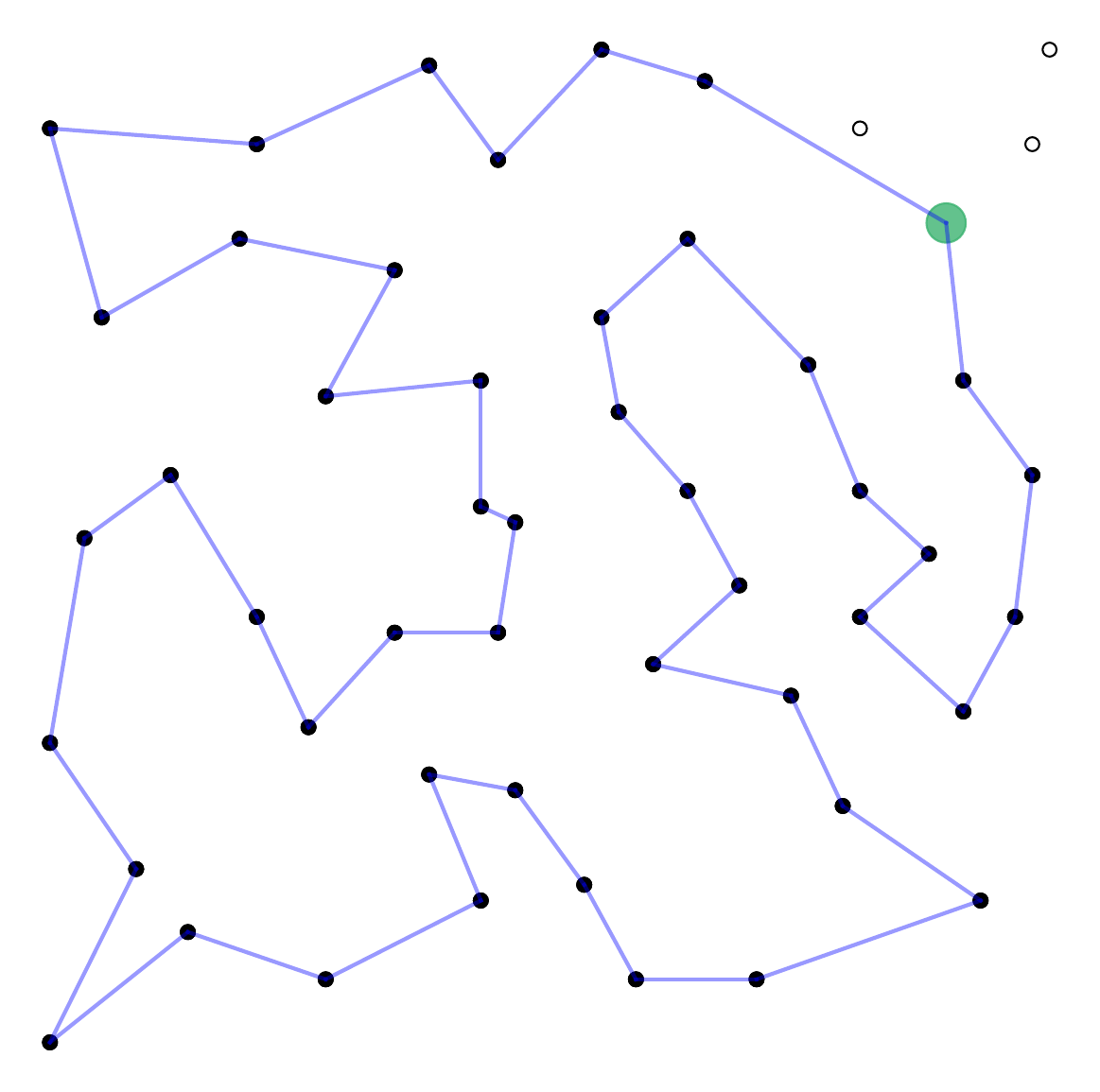}
		\text{Step 48}
	\end{minipage}
 	\begin{minipage}{0.19\linewidth}
		\centering
		\includegraphics[width=0.9\linewidth]{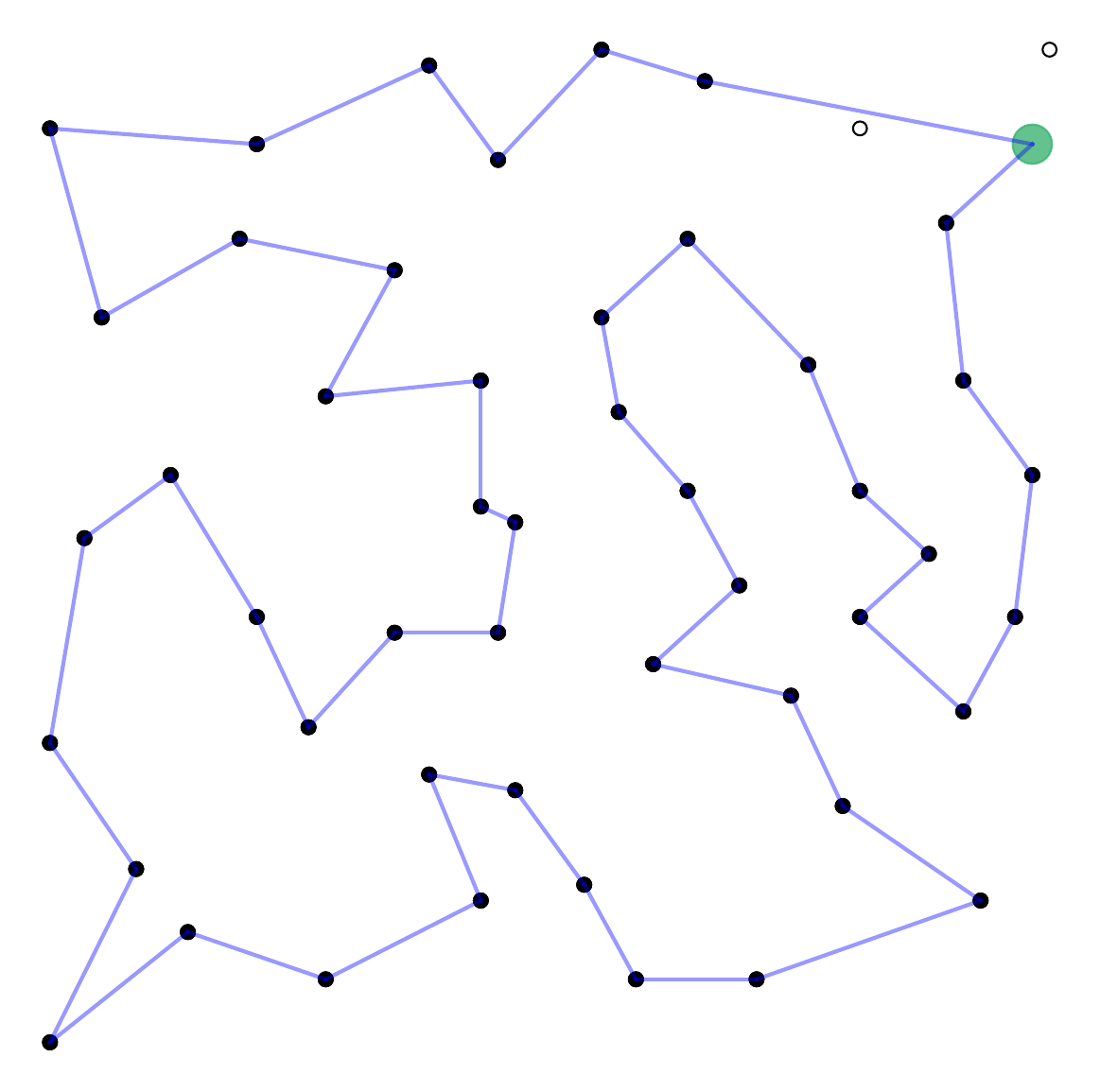}
		\text{Step 50}
	\end{minipage}
	\begin{minipage}{0.19\linewidth}
		\centering
		\includegraphics[width=0.9\linewidth]{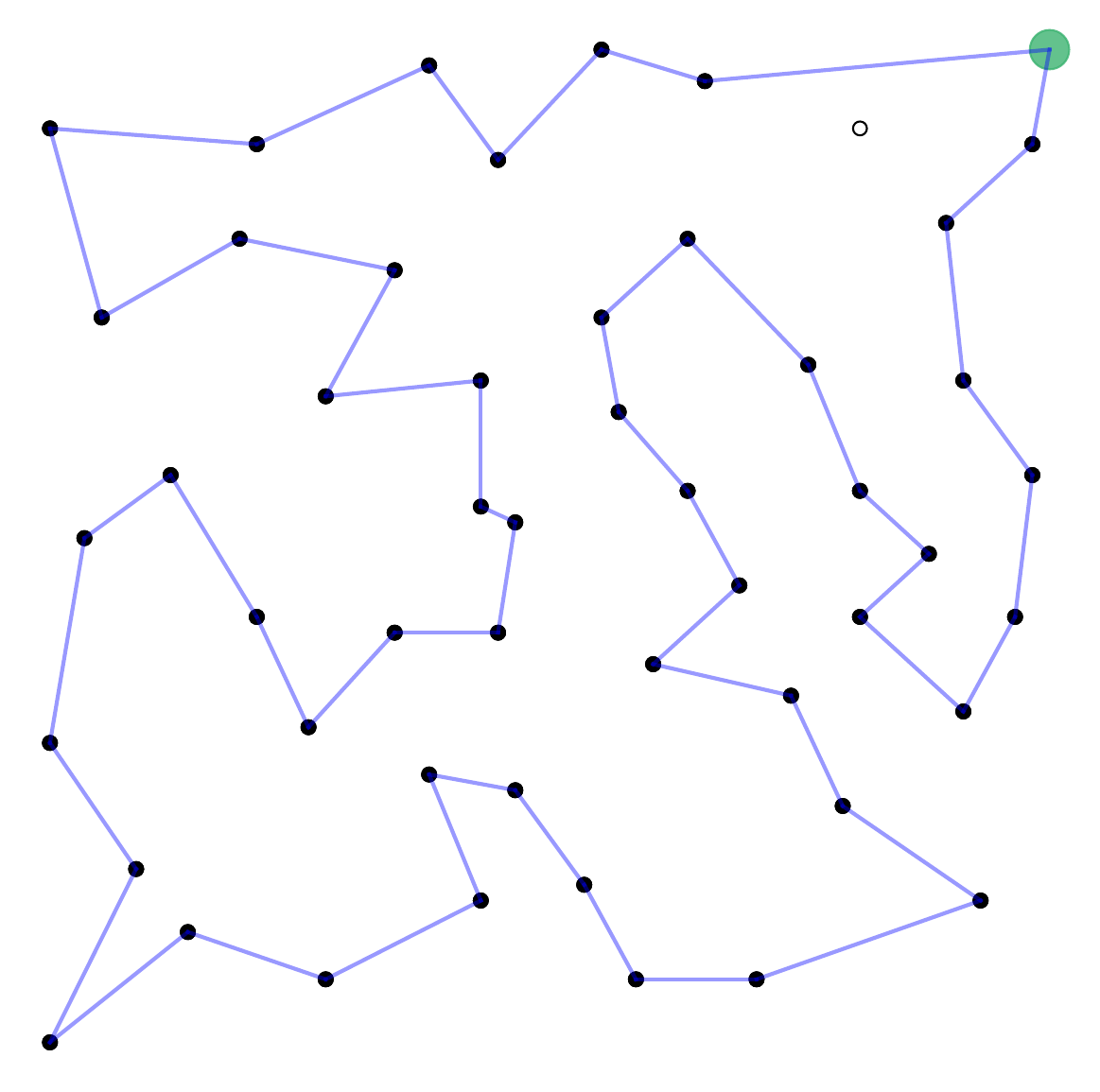}
		\text{Step 50}
	\end{minipage}
 	\begin{minipage}{0.19\linewidth}
		\centering
		\includegraphics[width=0.9\linewidth]{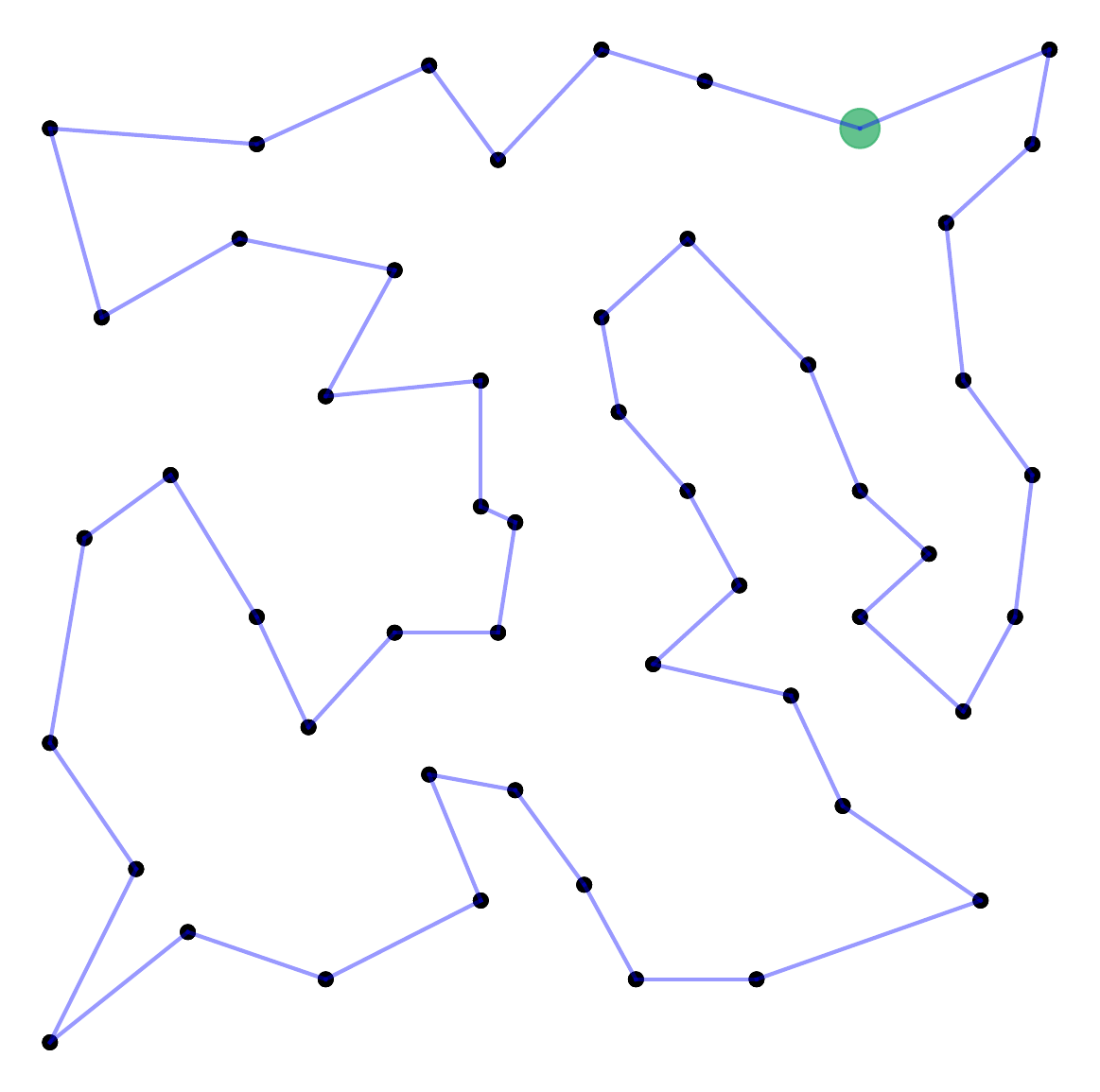}
		\text{Step 51}
	\end{minipage}
 \caption{Insertion-based construction process on the TSPLIB instance ``eil51''. The green dot is the node selected in the current step for insertion. The hollow dots are the unvisited nodes. The black dots are visited nodes.}
 \label{Fig:Insertion problem on TSP}
\end{figure}

\begin{figure}[htbp]
	\centering
	\begin{minipage}{0.19\linewidth}
		\centering
		\includegraphics[width=0.9\linewidth]{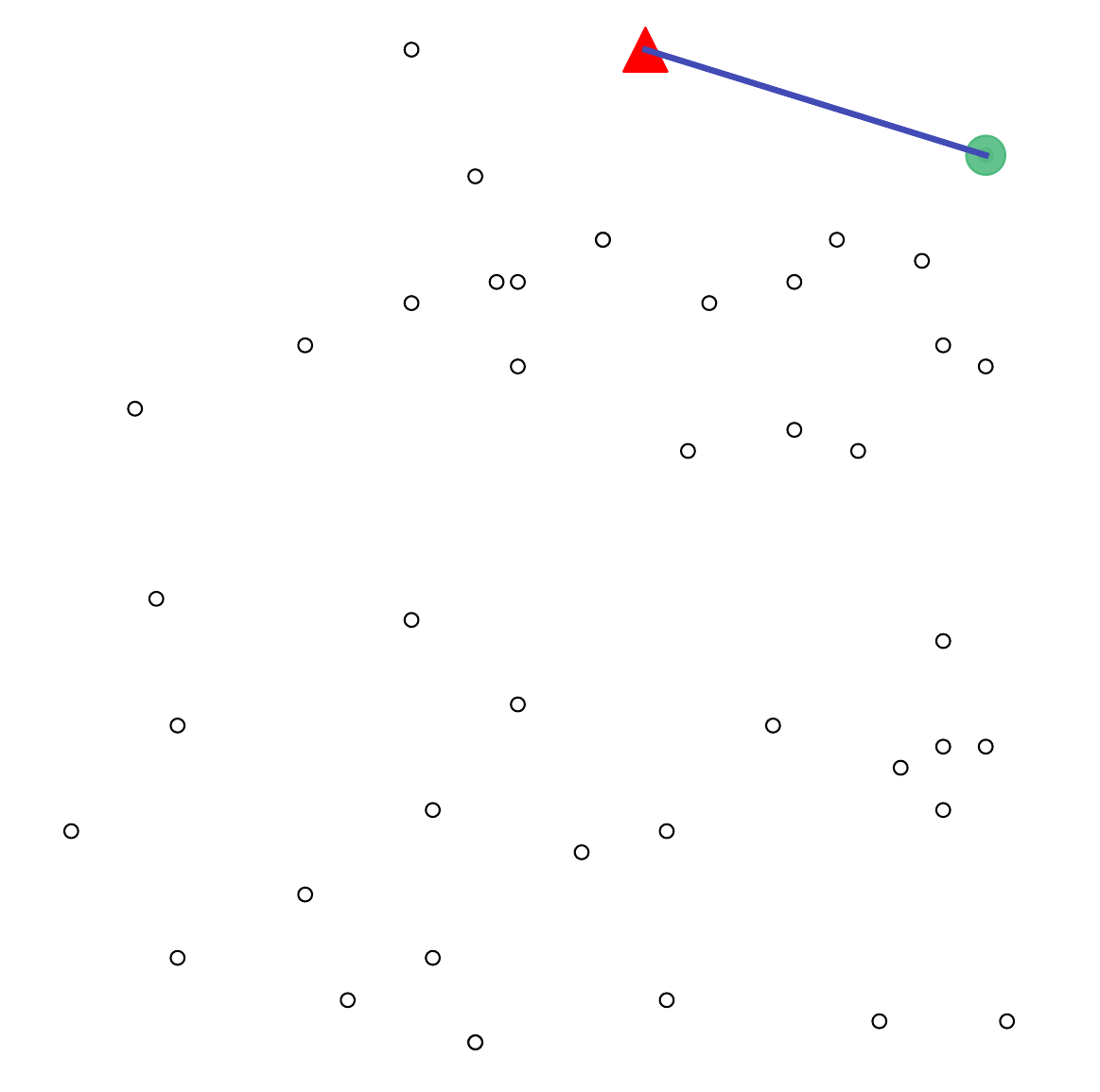}
		\text{Step 1}
	\end{minipage}
	\begin{minipage}{0.19\linewidth}
		\centering
		\includegraphics[width=0.9\linewidth]{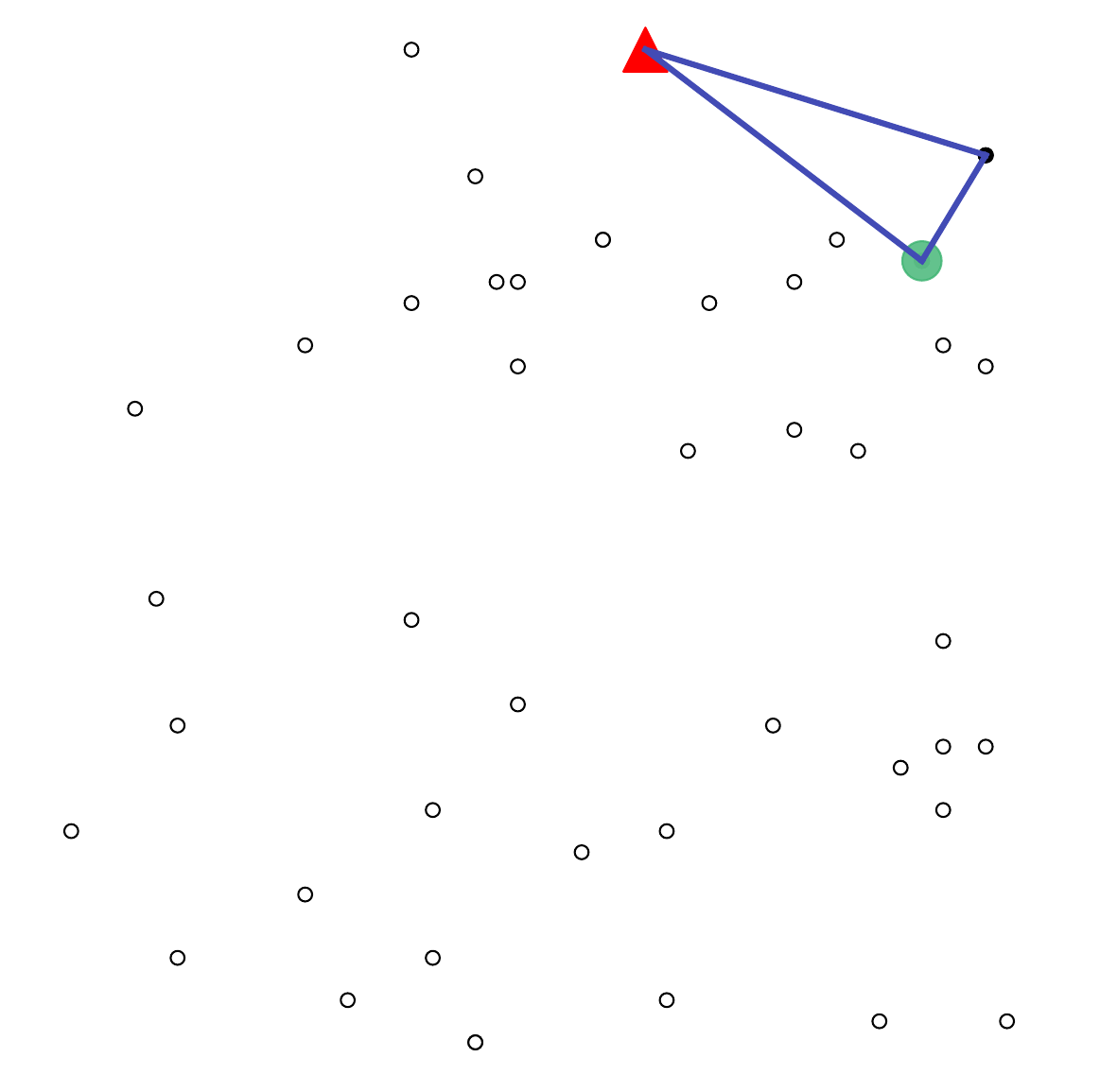}
		\text{Step 2}
	\end{minipage}
 	\begin{minipage}{0.19\linewidth}
		\centering
		\includegraphics[width=0.9\linewidth]{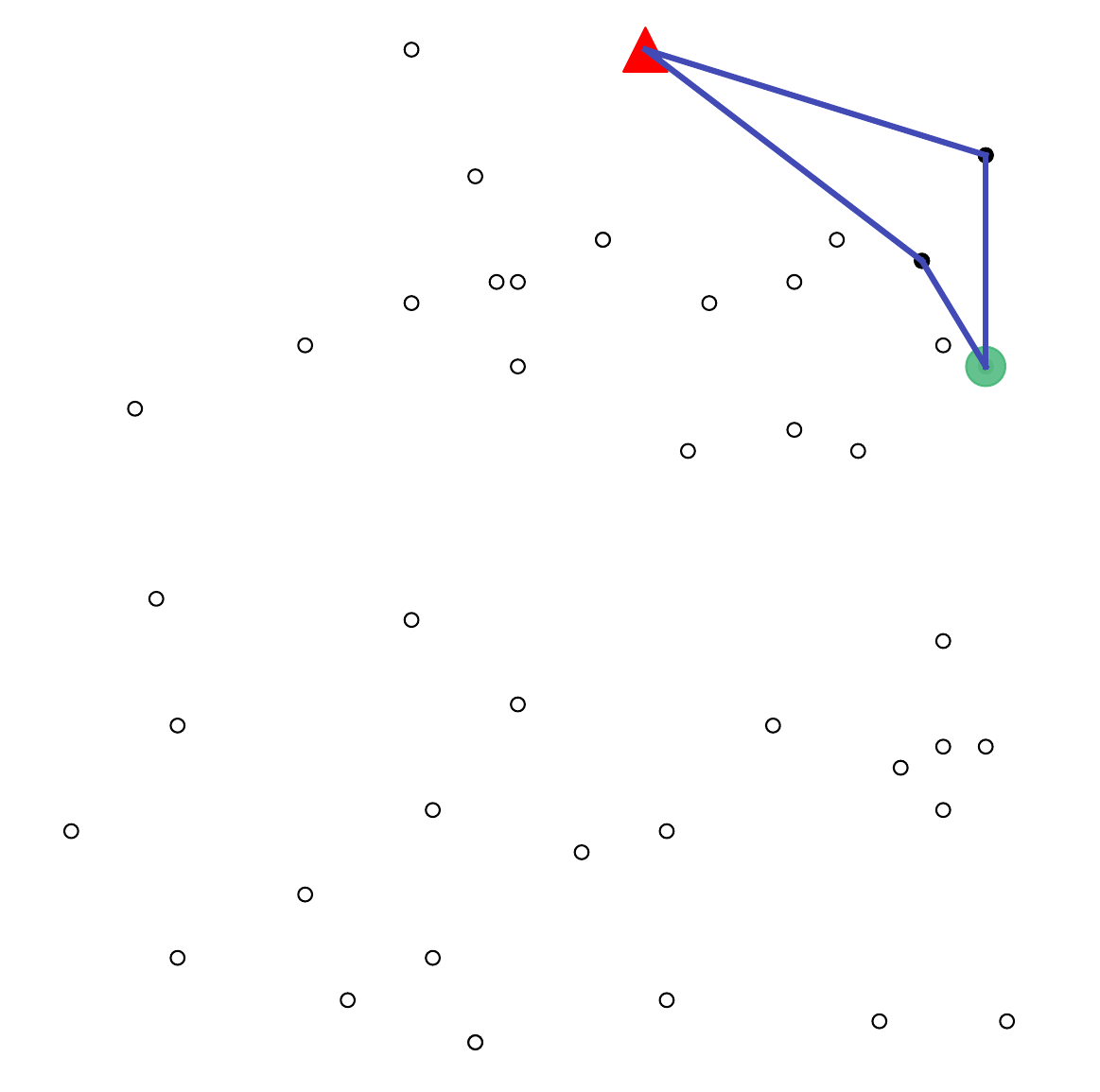}
		\text{Step 3}
	\end{minipage}
	\begin{minipage}{0.19\linewidth}
		\centering
        \text{......}
	\end{minipage}
 	\begin{minipage}{0.19\linewidth}
		\centering
		\includegraphics[width=0.9\linewidth]{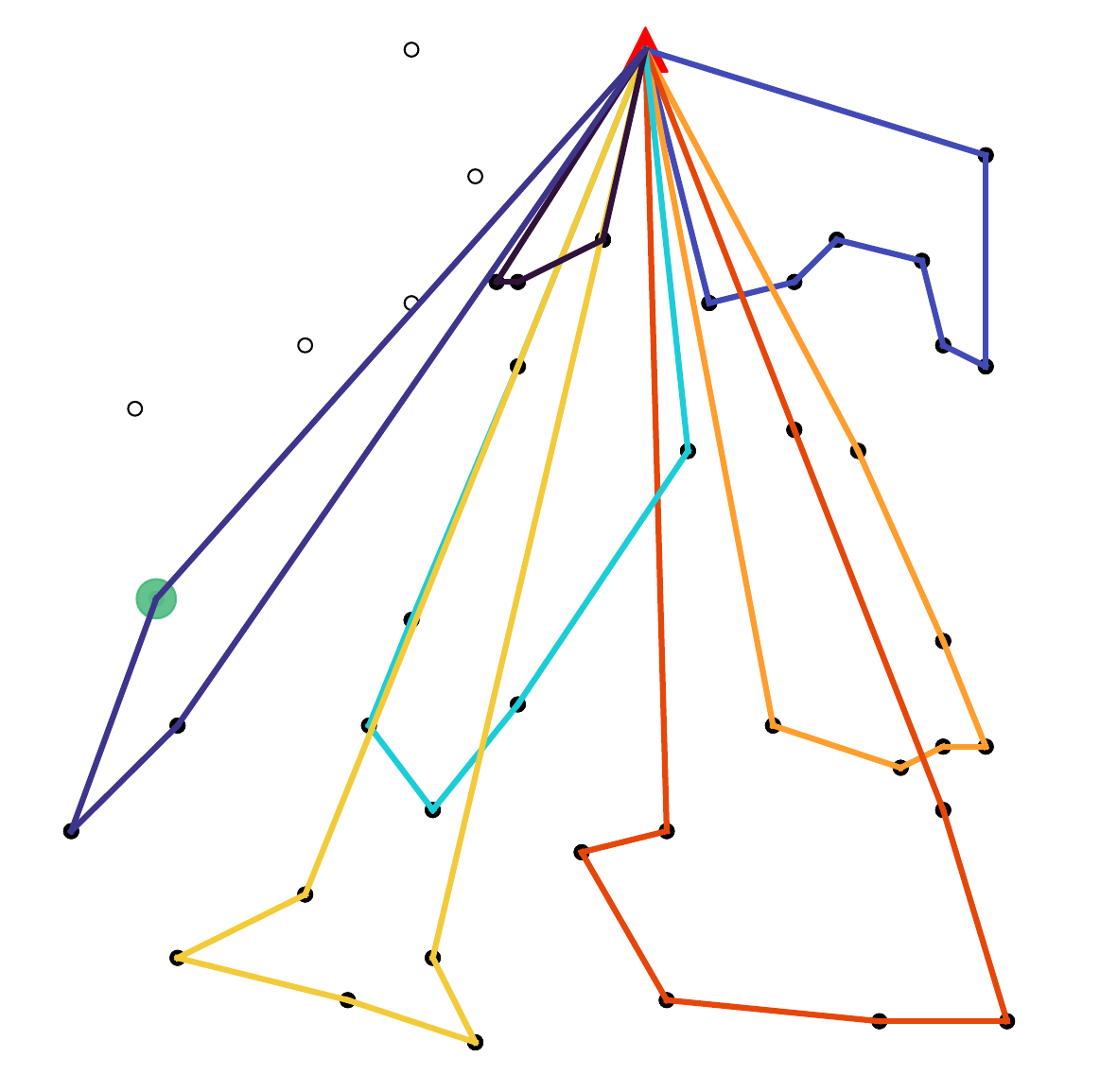}
		\text{Step 39}
	\end{minipage}
	\qquad

		\begin{minipage}{0.19\linewidth}
		\centering
		\includegraphics[width=0.9\linewidth]{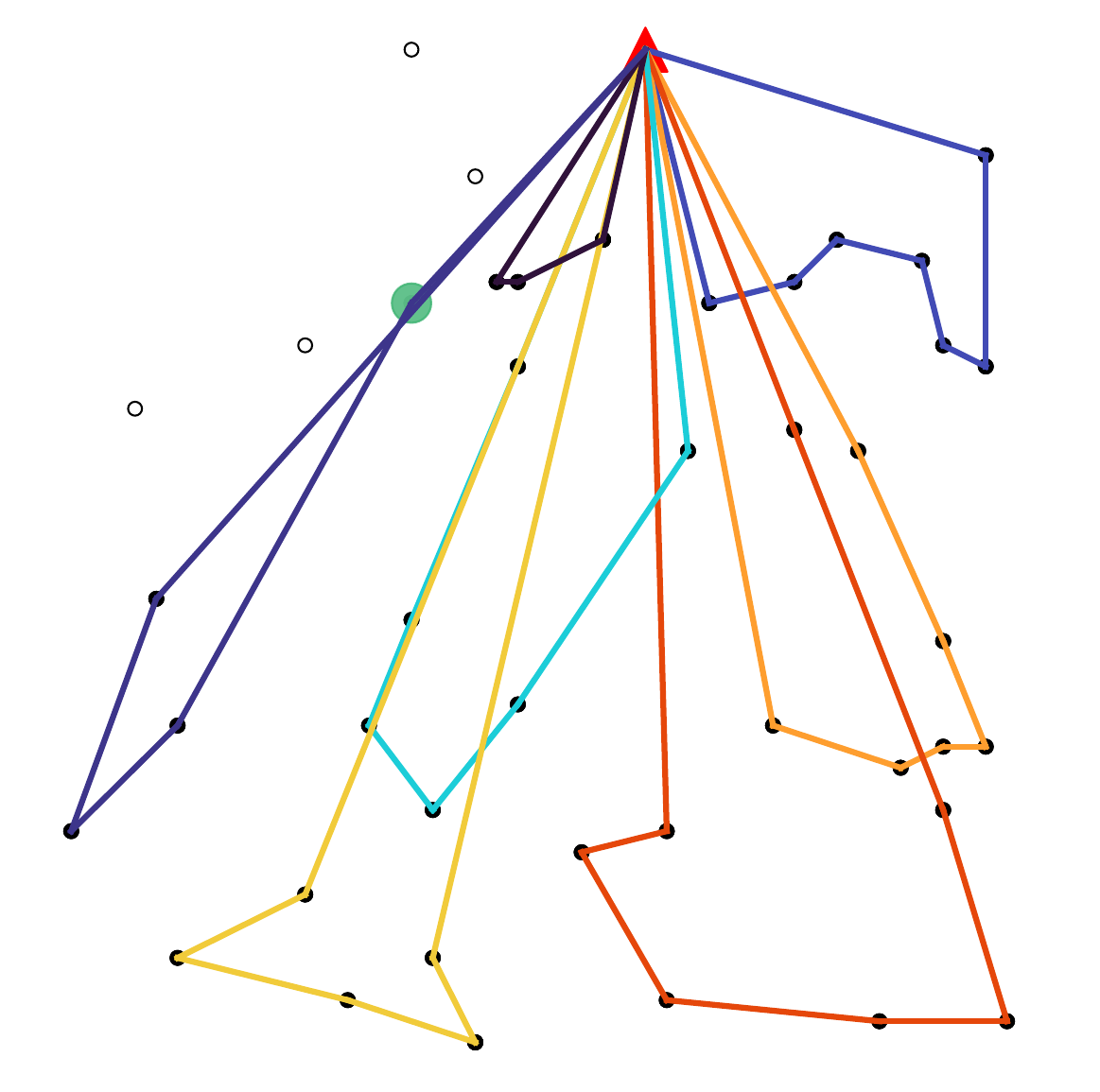}
		\text{Step 40}
	\end{minipage}
	\begin{minipage}{0.19\linewidth}
		\centering
		\includegraphics[width=0.9\linewidth]{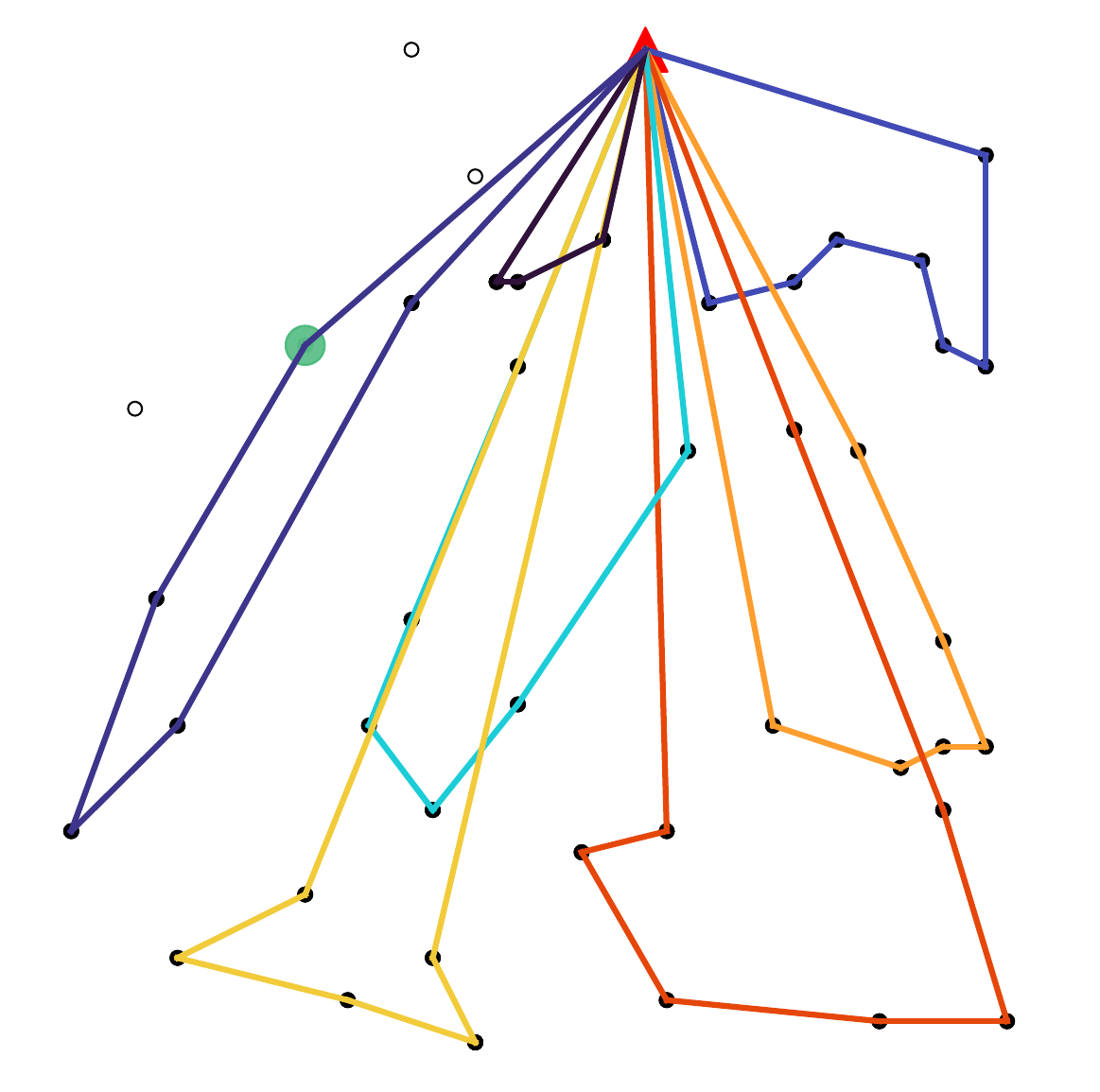}
		\text{Step 41}
	\end{minipage}
	\begin{minipage}{0.19\linewidth}
		\centering
			\includegraphics[width=0.9\linewidth]{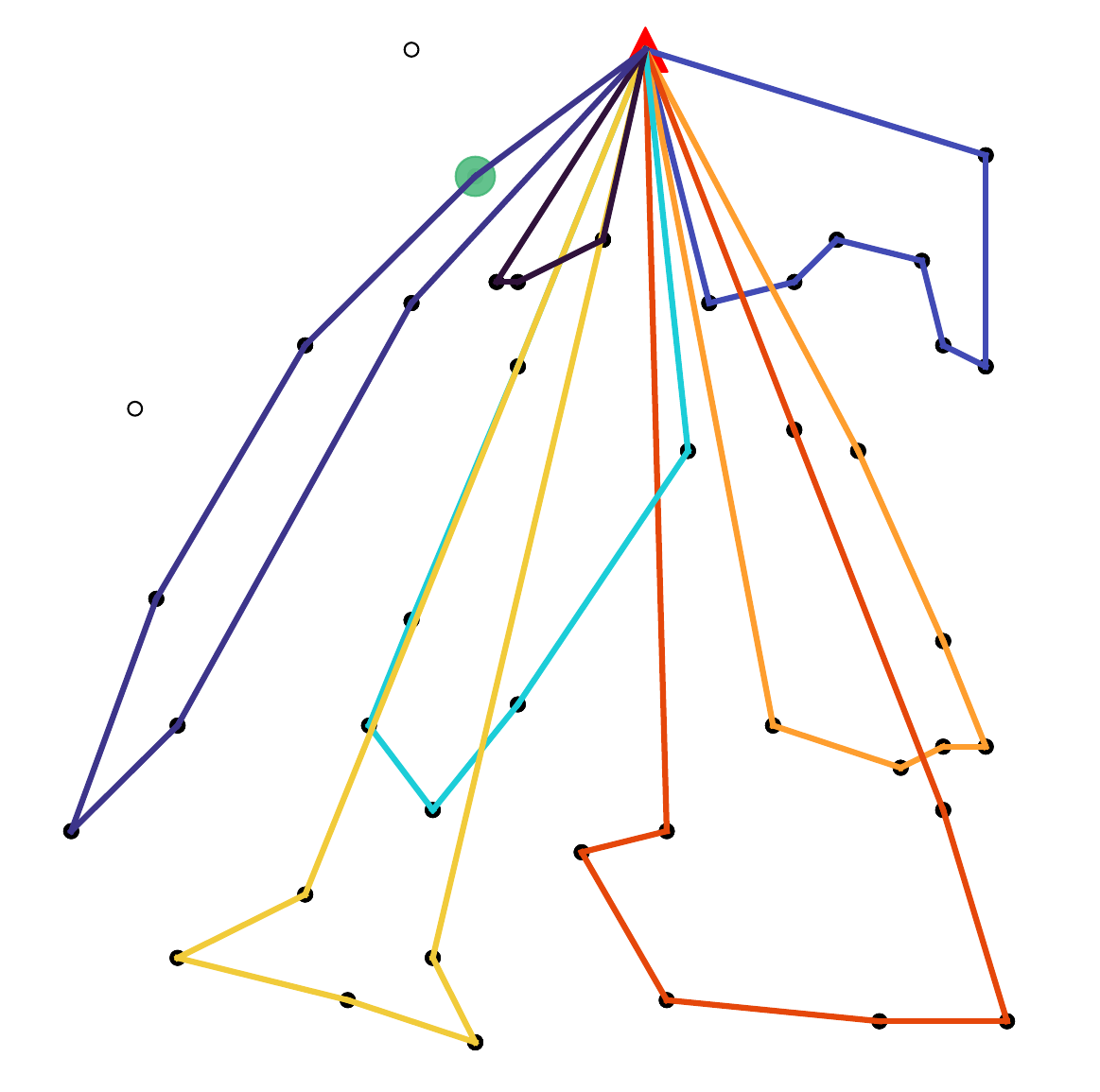}
		\text{Step 42}
	\end{minipage}
 	\begin{minipage}{0.19\linewidth}
		\centering
		\includegraphics[width=0.9\linewidth]{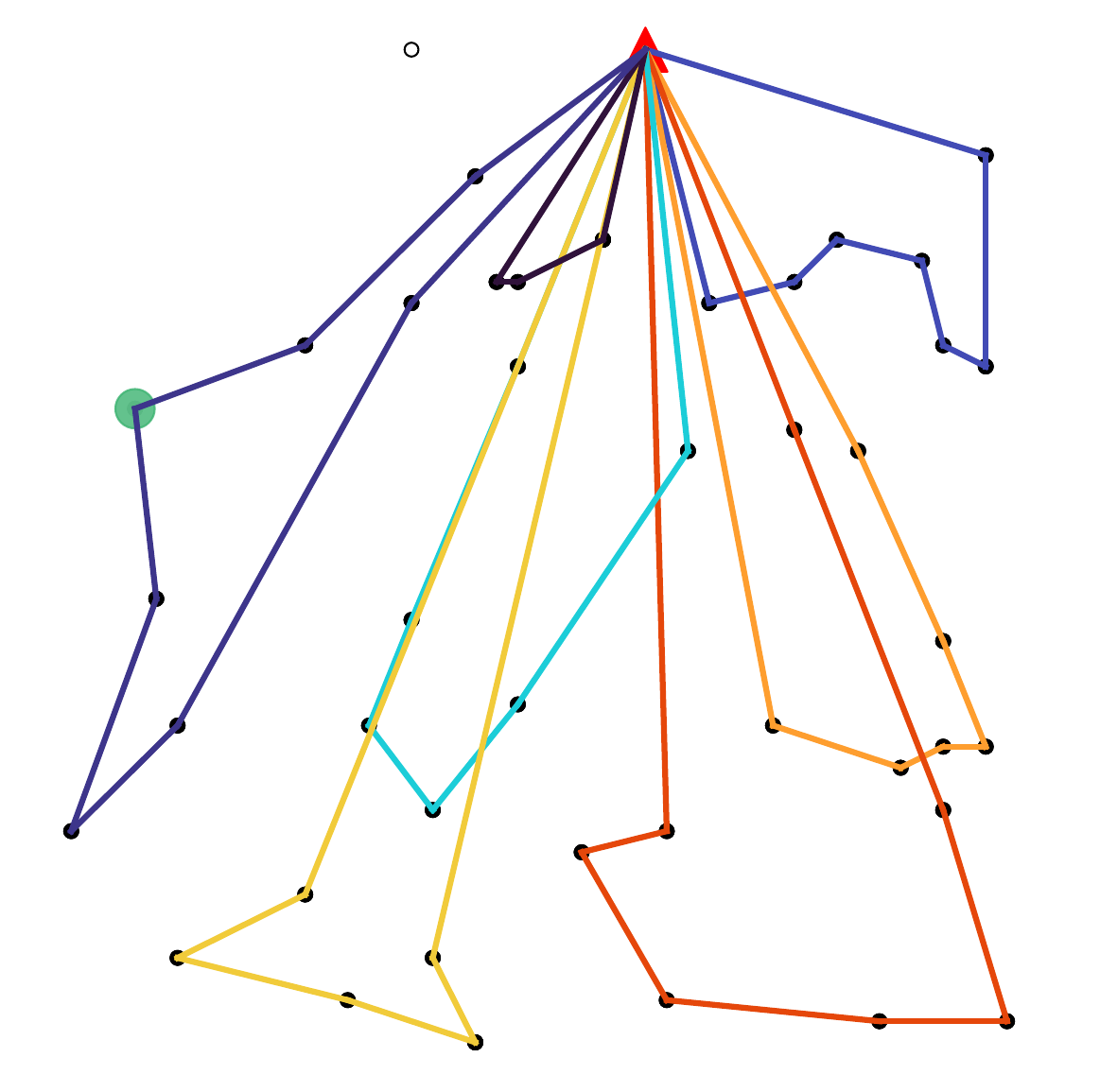}
		\text{Step 43}
	\end{minipage}
	\begin{minipage}{0.19\linewidth}
		\centering
		\includegraphics[width=0.9\linewidth]{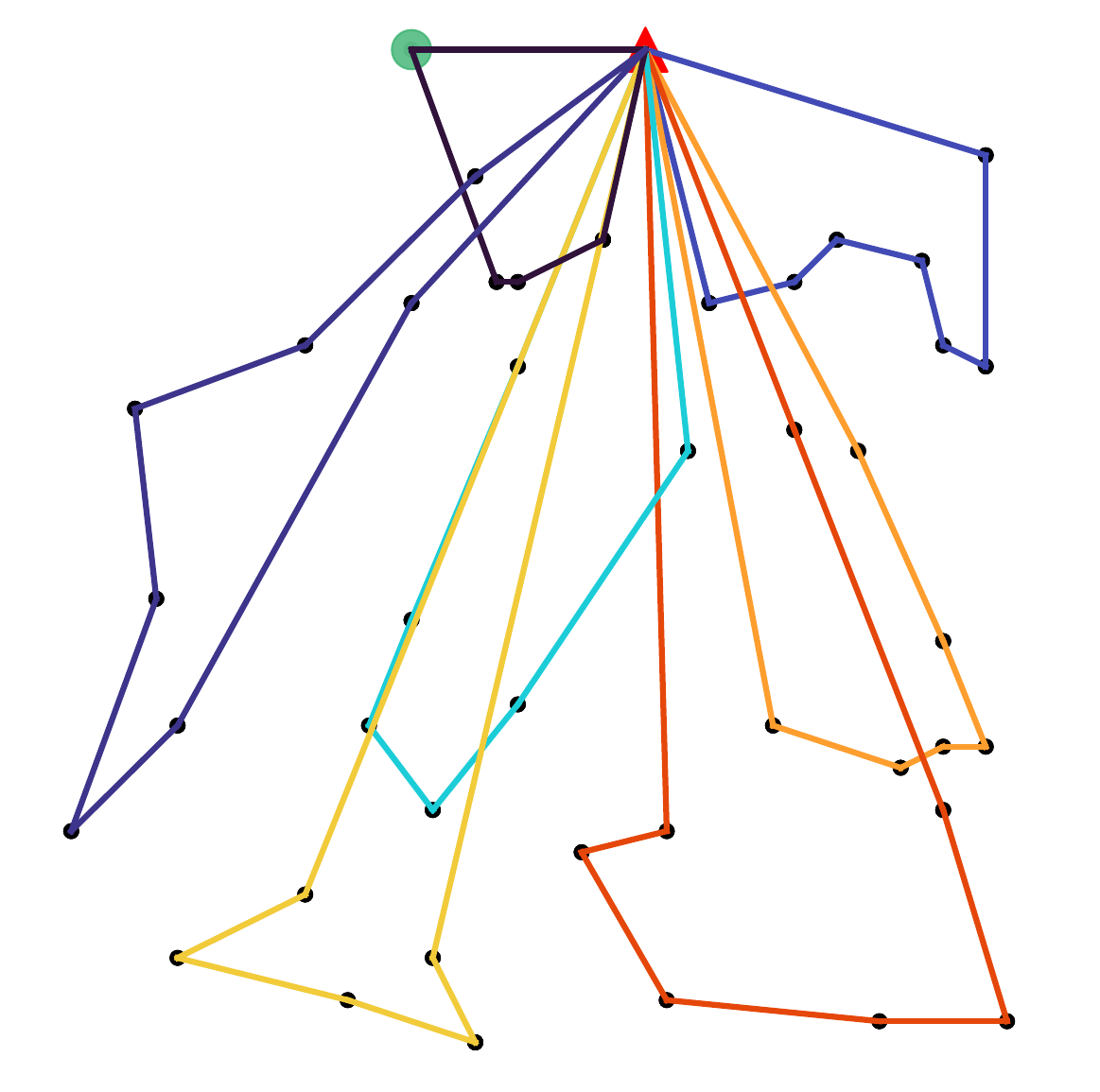}
		\text{Step 44}
	\end{minipage}
 \caption{Insertion-based construction process on the CVRPLIB instance ``A-n45-k7''. The red triangle is the depot node, the green dot is the unvisited customer node selected in the current step for insertion. The hollow dots are the unvisited nodes. The black dots are visited nodes.}
  \label{Fig:Insertion problem on CVRP}
\end{figure}

In this section, we provide graphical examples from both TSPLIB and CVRPLIB to illustrate the insertion-based construction process clearly. As shown in Figure ~\ref{Fig:Insertion problem on TSP}, the process begins by randomly selecting an unvisited node as the starting node for the incomplete solution. The model then continuously selects unvisited points to insert into the incomplete solution. Notably, in step 47, there are no unvisited nodes near the current node that can be connected without crossing the existing path. If the construction process is appending-based, the next step will involve directly connecting the last node to a distant unvisited node, resulting in a very long edge. However, the insertion-based construction allows for the direct insertion of the distant unvisited node into the incomplete solution (as seen in step 48),  thereby avoiding the generation of the long edge mentioned above. Similarly, in Figure~\ref{Fig:Insertion problem on CVRP}, steps 42 to 43 demonstrate a similar behavior of error correction in previous steps. This highlights the flexibility of the insertion-based construction.

\newpage

\section{Detailed Formulation of Attention Layer}
\label{append:AttnLayer}

\subsection{Attention Layer}
The attention layer in the transformer model~\citep{vaswani2017attention} is widely adopted in NCO for solving VRPs, mainly including a multi-head attention ($\mbox{MHA}$) sublayer and a node-wise feed-forward ($\mbox{FF}$) sublayer. Each sublayer is followed by a skip-connection~\cite{he2016deep} and normalization ($\mbox{Norm}$)~\citep{ioffe2015batch}. We denote $X^{(\ell-1)} \in \mathbb{R}^{n\times d}$ as the inputs of the $\ell$-th attention layer, the output of the attention layer is calculated as:
\begin{equation}
\begin{aligned}
     \hat{X}&=\mbox{Norm}(\mbox{MHA}(X^{(\ell-1)}) +X^{(\ell-1)}), \\
    X^{(\ell)}&=\mbox{Norm}(\mbox{FF}(\hat{X}) +\hat{X}).\\
\end{aligned}
\label{Attn}
\end{equation}  
We denote this calculation process as follows:
\begin{equation}
\begin{aligned}
     X^{(\ell)}&=\mbox{AttnLayer}(X^{(\ell-1)}).
\end{aligned}
\label{AttnLayer}
\end{equation} 
In this paper, the normalization is removed from the attention layer for enhancing generalization performance following~\citet{luo2023neural}. For readability, we omit the $(\ell)$ and $(\ell-1)$ in the following context.
\subsection{Multi-head Attention} We first describe the single-head self-attention function as follows:
\begin{equation}
\begin{aligned}
    \text{Attn}(X) &= \mbox{softmax}\left( \frac{ X W_Q (X W_K)^{\intercal}}{\sqrt{d}}\right)X W_V,
\end{aligned}
\label{attn}
\end{equation}
where $X \in \mathbb{R}^{n\times d}$ is the input matrice, and $W_Q\in \mathbb{R}^{d\times d_k}, W_K\in \mathbb{R}^{d\times d_k}, W_V \in \mathbb{R}^{d\times d_v}$ are learnable parameters. 

The multi-head attention sublayer applies the single-head attention function in Equation (\ref{attn}) for $h$ times in parallel with independent parameters: 
\begin{equation}
\begin{aligned}
    \mbox{MHA}(X) &= \mbox{Concat}(\mbox{head}_1,\ldots,\mbox{head}_h) W^O, \\
     \mbox{head}_i&=\mbox{Attn}_i(X),
\end{aligned}
\label{mha}
\end{equation}
where for each of $\mbox{Attn}_i(X)$, $d_k =  d_v = d/h$. $W^O \in \mathbb{R}^{d\times d}$ is a learnable matrix.

\subsection{Feed Forward Layer}
\begin{equation}
\begin{aligned}
     \mbox{FF}(X) = \mbox{max}(0, XW_{f1}+b_{f1})W_{f2}+b_{f2}, 
\end{aligned}
\end{equation}
where $W_{f1} \in \mathbb{R}^{d\times d_{ff}}$,  $b_{f1} \in \mathbb{R}^{d_{ff}}  $ and $W_{f2} \in \mathbb{R}^{d_{ff}\times d}, b_{f2} \in \mathbb{R}^{d}$ are learnable parameters.

\newpage

\section{Broader Impacts}
\label{Broader Impacts}
This research advances the field of neural combinatorial optimization by introducing the L2C-Insert framework, which employs advanced machine learning techniques to solve Vehicle Routing Problems (VRPs). We believe this framework will provide valuable insights and inspire subsequent research focused on developing more efficient and effective neural methods for VRPs. Moreover, as a general learning-based approach for solving VRPs, the proposed L2C-Insert framework is not anticipated to have specific negative societal impacts.

\section{Licenses}
\label{Licenses}
The licenses for the codes and the datasets used in this work are listed in Table \ref{appendix: Licenses}.

\begin{table}[htbp]
\centering
\caption{List of licenses for the codes and datasets we used in this work}
\label{appendix: Licenses}
\resizebox{0.99\textwidth}{!}{%
\begin{tabular}{l |l | l | l }
\toprule
 Resource   &   Type  &  Link  & License    \\
\midrule
LKH3 \citep{LKH3} & Code & http://webhotel4.ruc.dk/~keld/research/LKH-3/ & Available for academic research use\\
HGS \citep{HGS}& Code & https://github.com/chkwon/PyHygese & MIT License\\
Concorde \citep{applegate2006concorde}& Code & https://github.com/jvkersch/pyconcorde &  BSD 3-Clause License \\
POMO \citep{kwon2020pomo}& Code & https://github.com/yd-kwon/POMO & MIT License \\
Att-GCN+MCTS \citep{fu2021generalize} & Code & https://github.com/SaneLYX/TSP\_Att-GCRN-MCTS & MIT License \\
LEHD~\citep{luo2023neural} & Code & https://github.com/CIAM-Group/NCO\_code/tree/main/ & MIT License \\
 &  & single\_objective/LEHD &  \\
BQ~\citep{drakulic2023bq}  & Code & https://github.com/naver/bq-nco & CC BY-NC-SA 4.0 \\
GLOP~\citep{ye2024glop} & Code & https://github.com/henry-yeh/GLOP & MIT License  \\
HTSP~\citep{pan2023h-tsp} & Code & https://github.com/Learning4Optimization-HUST/H-TSP & MIT License \\
DIMES~\citep{qiu2022dimes} & Code & https://github.com/DIMESTeam/DIMES & MIT License \\
DIFUSCO~\citep{sun2023difusco} & Code & https://github.com/Edward-Sun/DIFUSCO  & MIT License \\
UDC~\citep{zheng2024udc} & Code & https://github.com/CIAM-Group/NCO\_code/tree/main/ & MIT License \\
&  & single\_objective/UDC-Large-scale-CO-master & \\
ELG~\citep{gao2024towards} & Code & https://github.com/gaocrr/ELG &  MIT License \\
INViT~\citep{fang2024invit} & Code & https://github.com/Kasumigaoka-Utaha/INViT & Available for academic research use \\
NeurOpt~\citep{ma2023neuopt} & Code & https://github.com/yining043/NeuOpt & MIT License  \\
\midrule
TSPLib~\citep{Rein91} & Dataset & http://comopt.ifi.uni-heidelberg.de/software/TSPLIB95/ & Available for any non-commercial use \\
CVRPLib~\citep{uchoa2017new} & Dataset & http://vrp.galgos.inf.puc-rio.br/index.php/en/  & Available for academic research use \\
\bottomrule
\end{tabular}%
}
\end{table}

\clearpage

\newpage

\end{CJK*}
\end{document}